\documentclass[a4paper,fleqn]{cas-dc}

\usepackage[authoryear,sort]{natbib}
\usepackage{lastpage}

\usepackage{bibunits}
\defaultbibliographystyle{cas-model2-names}
\defaultbibliography{references}

\ExplSyntaxOn

\cs_new:Npn \ResetForAppendixFrontMatterI
 {
   \seq_gclear:N \g_stm_title_seq
   \seq_gclear:N \g_stm_prelimsau_seq
   \seq_gclear:c { g_stm_au0_seq }
   \seq_gclear:c { g_stm_clau0_seq }
   \seq_gclear:c { g_stm_aff0_seq }
   \seq_gclear:N \g_stm_fnote_seq
   \seq_gclear:N \g_stm_cor_seq
   \seq_gclear:N \g_stm_tnote_seq
   \seq_gclear:N \g_stm_nonumnote_seq
   \seq_gclear:N \g_stm_maltese_seq
   \int_gzero:N \g_stm_fnote_int
   \int_gzero:N \g_stm_cor_int
   \int_gzero:N \g_stm_tnote_int
   \setcounter{footnote}{0}
   \setcounter{section}{0}
   \setcounter{figure}{0}
   \setcounter{table}{0}
   \setcounter{equation}{0}
   \int_gset:Nn \g_stage_int { 1 }
 }

\int_new:N \g_stage_int
\int_gzero:N \g_stage_int

\seq_new:N \g_scratch_seq
\cs_new:Nn \__append_orcid_item:n
 { \seq_gput_right:Nx \g_scratch_seq { \seq_item:Nn \g_stm_orcid_seq {#1} } }
\cs_new:Nn \__append_ead_item:n
 { \seq_gput_right:Nx \g_scratch_seq { \seq_item:Nn \g_stm_ead_seq {#1} } }

\cs_new:Nn \PrintOrcidRange:nn
 {
   \seq_gclear:N \g_scratch_seq
   \int_step_function:nnnN {#1} {1} {#2} \__append_orcid_item:n
   \seq_if_empty:NF \g_scratch_seq
   {
     \group_begin:
     \tex_let:D \thefootnote \relax
     \footnotetext
     { \raggedright \textsc{orcid}(s):\c_space_token \seq_use:Nn \g_scratch_seq {;~} }
     \group_end:
   }
 }

\cs_new:Nn \PrintEmailRange:nn
 {
   \seq_gclear:N \g_scratch_seq
   \int_step_function:nnnN {#1} {1} {#2} \__append_ead_item:n
   \seq_if_empty:NF \g_scratch_seq
   {
     \group_begin:
     \tex_let:D \thefootnote \relax
     \footnotetext
     { \raggedright \includegraphics[height=8pt]{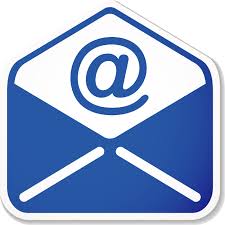}\c_space_token \seq_use:Nn \g_scratch_seq {;~} }
     \group_end:
   }
 }

\cs_set:Npn \printorcid
 { \int_compare:nNnTF \g_stage_int = {0} { \PrintOrcidRange:nn{1}{5} } { \PrintOrcidRange:nn{6}{10} } }
\cs_set:Npn \printemails
 { \int_compare:nNnTF \g_stage_int = {0} { \PrintEmailRange:nn{1}{1} } { \PrintEmailRange:nn{2}{2} } }
\ExplSyntaxOff

\newcommand \ResetForAppendix { \ResetForAppendixFrontMatterI }

\begin{document}
\csdef{lastpage}{\pageref{LastPage}}
\let\WriteBookmarks\relax
\def\floatpagepagefraction{1}
\def\textpagefraction{.001}

\begin{bibunit}
\csdef{lastpage}{\pageref{LastPage}}
\let\WriteBookmarks\relax
\def\floatpagepagefraction{1}
\def\textpagefraction{.001}


\shorttitle{Multi-Level Evidence Aggregation for Facial Phenotype Retrieval}
\shortauthors{A. Hustinx, C. Kaffin\'e et~al.}

\title[mode = title]{Multi-Level Evidence Aggregation for Robust Facial Phenotype Retrieval in Rare Genetic Disorder Prioritization}

\author[1]{Alexander Hustinx}[orcid=0000-0003-4592-3979]
\fnmark[1]
\cormark[1]
\ead{ahustinx@uni-bonn.de}
\credit{Conceptualization, Methodology, Software, Formal analysis, Investigation, Data curation, Validation, Visualization, Supervision, Project administration, Writing - original draft, Writing - review \& editing}

\author[1]{Carolin Kaffin\'e}[orcid=0009-0003-3896-6847]
\fnmark[1]
\credit{Conceptualization, Methodology, Software, Formal analysis, Investigation, Data curation, Validation, Visualization, Writing - original draft, Writing - review \& editing}

\author[1]{Behnam Javanmardi}[orcid=0000-0002-9317-6114]
\credit{Methodology, Validation, Supervision, Writing - review \& editing}

\author[1]{Tzung-Chien Hsieh}[orcid=0000-0003-3828-4419]
\credit{Conceptualization, Methodology, Resources, Data curation, Supervision, Project administration, Writing - original draft, Writing - review \& editing}

\author[1]{Peter Krawitz}[orcid=0000-0002-3194-8625]
\credit{Methodology, Resources, Supervision, Writing - review \& editing}

\affiliation[1]{organization={Institute for Genomic Statistics and Bioinformatics, University Hospital Bonn},
                city={Bonn},
                country={Germany}}

\fntext[1]{Authors contributed equally.}
\cortext[1]{Corresponding author.}

\begin{abstract}
AI-assisted facial phenotyping supports rare genetic disorder prioritization by retrieving visually similar diagnosed cases from facial image reference databases such as the GestaltMatcher Database (GMDB). Existing GestaltMatcher-based retrieval frameworks compare each test image with individual gallery images in a facial phenotype embedding space. However, this pointwise formulation does not fully exploit available evidence, because patients may have multiple images and disorders may be represented by multiple diagnosed gallery patients.

We propose an inference-time multi-level evidence aggregation framework that improves facial phenotype retrieval without modifying the underlying GestaltMatcher-Arc encoder. The framework combines embedding-level patient aggregation of multiple images from the same individual, patient-weighted disorder centroids, and hybrid individual-centroid scoring to integrate test-patient observations, disorder-level gallery evidence, and local nearest-neighbor evidence.

We evaluated the approach on GMDB v1.1.4 across disorders represented during training (GMDB-Freq), unseen disorders (GMDB-Rare), and multi-image patient subsets, using a unified gallery containing both GMDB-Freq and GMDB-Rare disorders. Multi-level evidence aggregation improved mean per-disorder top-$N$ retrieval accuracy across all evaluation subsets. Top-1 accuracy increased from 38.52\% to 48.82\% on GMDB-Freq and from 19.38\% to 23.79\% on GMDB-Rare. On multi-image subsets, top-1 accuracy increased from 46.12\% to 60.94\% on GMDB-Multi-Freq and from 18.54\% to 26.71\% on GMDB-Multi-Rare.

These findings show that inference-time aggregation can improve next-generation facial phenotype retrieval without retraining the encoder, supporting a shift from isolated single-image matching toward multi-level aggregation of patient and disorder evidence for rare-disorder prioritization.
\end{abstract}

\begin{keywords}
rare genetic disorders \sep facial phenotyping \sep medical image retrieval \sep evidence aggregation
\end{keywords}

\maketitle


\section{Introduction}\label{sec:intro}

Rare diseases are individually uncommon but collectively affect an estimated 3.5-5.9\% of the global population, corresponding to hundreds of millions of people worldwide \citep{nguengangwakap2020}. Diagnosis remains challenging because individual disorders are rare, clinical and facial manifestations can vary between affected individuals, and recognition often requires specialized clinical expertise. Around 30–40\% of genetic disorders are associated with recognizable facial phenotypes, and facial gestalt can provide an important diagnostic clue during clinical evaluation \citep{ferry2014}. However, recognizing subtle or rare facial patterns is difficult, particularly when few diagnosed reference cases are available or when disorders have age-dependent facial phenotypes, the patient is of an underrepresented ancestry, or imaging is performed in non-ideal conditions.

AI-assisted facial phenotyping systems aim to support this process by comparing a patient's facial image with reference cases of patients with known genetic disorders. GestaltMatcher and GestaltMatcher-Arc (GM-Arc) formulate this task as retrieval in a facial phenotype embedding space, the Clinical Face Phenotype Space (CFPS). In the CFPS, facial images are embedded and ranked according to similarity to diagnosed gallery images \citep{hsieh2022,hustinx2023}. This retrieval-based formulation is particularly relevant for ultra-rare disorders, where too few diagnosed cases exist to train a disorder-specific classifier but a comparison against individual reference patients remains possible.

Despite these advances, current retrieval workflows largely treat each facial image independently and each gallery image as an independent reference point. This point-wise formulation does not fully exploit the structure of the available evidence. A patient may be represented by multiple images that capture different aspects of the facial phenotype across ages, head poses, expressions, or imaging conditions. Similarly, a disorder is represented not by a single canonical face, but by a collection of diagnosed individuals that may sample only part of the range of facial presentations associated with that disorder. As a result, retrieval based on isolated image-level matches can be sensitive to suboptimal test images, outlier gallery images, and imbalanced gallery representations.

\begin{figure*}[pos=!ht]
  \centering
  \includegraphics[width=\linewidth]{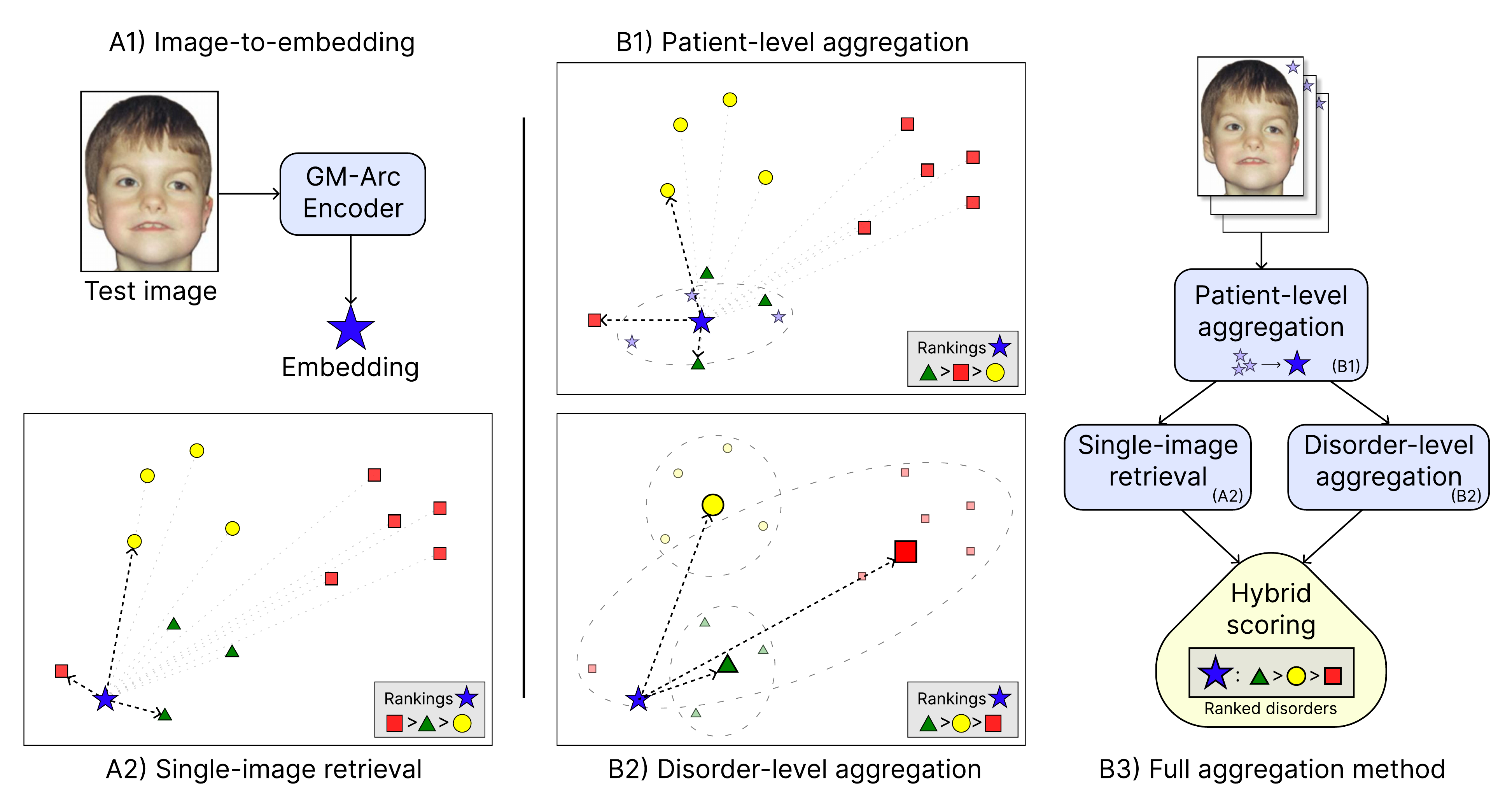}
  \caption{Conceptual overview of multi-level evidence aggregation for rare-disorder retrieval. (A) Baseline GM-Arc retrieval. (A1) A synthetic test image* is embedded into the Clinical Face Phenotype Space. (A2) The resulting embedding is compared with individual gallery images from patients with confirmed diagnoses, and disorders are ranked according to the nearest gallery match. (B) Proposed inference-time aggregation strategies. (B1) Patient-level aggregation combines multiple test images from the same individual before disorder ranking. (B2) Disorder-level gallery aggregation summarizes gallery evidence using patient-weighted disorder centroids, denoted by the larger shapes. (B3) The full aggregation method combines patient-level aggregation with a hybrid individual-centroid scoring to produce a single ranked list of candidate disorders. The blue star denotes the test image; other colors/shapes denote different disorders. *) The facial image shows a synthetic rare-disorder-like face generated by GestaltGAN \citep{kirchhoff2025} for illustration purposes}
  \label{fig:overview}
\end{figure*}

We therefore investigate whether rare-disorder facial phenotype retrieval can be improved by aggregating evidence at inference time, without retraining or modifying the underlying facial phenotype encoder. Rather than treating patients and disorders as isolated points in the CFPS, we treat them as sets of observations. Multiple images of the same patient can provide complementary views of an individual phenotype, while multiple diagnosed gallery patients provide disorder-level evidence about the embedding region occupied by a disorder. Figure~\ref{fig:overview} provides a conceptual overview of this shift. Instead of treating each test image and each gallery image as independent points in the CFPS, the proposed framework aggregates evidence at the levels at which it is available: across multiple images of the same patient, across diagnosed patients belonging to the same disorder, and across local and global disorder scores. The figure illustrates the overall information flow, while the individual aggregation components are defined in Section~\ref{sec:methods}.

In this work, we propose an inference-time multi-level evidence aggregation framework for AI-assisted facial phenotype retrieval. The framework combines patient-level aggregation of multiple test images, patient-weighted disorder centroids for gallery aggregation, and hybrid individual-centroid scoring that integrates local nearest-neighbor evidence with global disorder-level evidence. We evaluate the framework on GMDB v1.1.4 using the GM-Arc retrieval pipeline across disorders represented during training, unseen rare disorders, and multi-image patient subsets. By evaluating the framework using a unified gallery containing both seen and unseen disorders, we assess retrieval in a broader and more heterogeneous reference setting. The proposed framework improves retrieval without retraining or otherwise modifying the underlying encoder, supporting a shift from isolated single-image matching toward multi-level aggregation of patient and disorder evidence.

\section{Related work}\label{sec:related}

The proposed framework builds on AI-assisted facial phenotyping for rare genetic disorders and on machine-learning methods for aggregating evidence across sets, prototypes, and exemplars. We first summarize facial phenotyping and retrieval-based rare-disorder prioritization, then discuss aggregation strategies that motivate the proposed inference-time framework. The focus is mainly on work directly related to retrieval and evidence aggregation, rather than on rare-disease diagnosis or medical image analysis more broadly.

\subsection{AI-assisted facial phenotyping and retrieval-based rare-disorder prioritization}\label{sec:related-phenotyping}

Facial gestalt has long been used as a diagnostic clue in clinical genetics, as many genetic disorders are associated with recognizable craniofacial patterns. Early computational work showed that ordinary facial photographs contain diagnostically relevant phenotype information and can be embedded in a Clinical Face Phenotype Space (CFPS), where patients with similar syndromic facial features tend to be located closer together \citep{ferry2014}. Deep-learning approaches subsequently demonstrated that facial images can support syndrome prioritization at larger scale. In particular, DeepGestalt used convolutional neural networks to classify hundreds of genetic syndromes from facial photographs, establishing AI-assisted facial phenotyping as a clinically relevant diagnostic-support technology \citep{gurovich2019}.

Retrieval-based approaches address an important limitation of closed-set classification in rare-disease diagnosis. Because many genetic disorders have few available training samples, disorder-specific classifiers may be difficult to train reliably. GestaltMatcher therefore formulated facial phenotyping as retrieval in CFPS, allowing a test image to be compared with diagnosed gallery images and enabling candidate-disorder ranking even for ultra-rare disorders with limited reference data \citep{hsieh2022}. GestaltMatcher-Arc further improved the facial phenotype representation by building on ArcFace-style angular-margin face representation learning and combining this with model ensembling and test-time augmentation for rare- and ultra-rare disorder verification \citep{deng2019,hustinx2023}. In parallel, the expansion of GMDB has increased the number of available images, patients, disorders, and metadata, enabling analyses of ancestry-related performance differences and the effect of gallery-set expansion \citep{lesmann2024}.

Despite these advances, standard retrieval workflows remain largely image-centric. Test images are typically evaluated independently, and gallery images are treated as independent reference points. For patients with multiple available images, this does not fully exploit complementary observations across age, head pose, expression, or acquisition conditions. Likewise, for gallery disorders represented by multiple patients and images, nearest-neighbor retrieval does not explicitly summarize disorder-level gallery structure and can be sensitive to isolated local matches or imbalance in the number of images per gallery patient. This leaves open how existing facial phenotype embeddings can be aggregated at inference time to improve disorder retrieval without retraining the encoder.

\subsection{Evidence aggregation, prototypes, and gallery-level representations}\label{sec:related-aggregation}

Machine-learning methods for set-structured data provide a useful conceptual basis for patient- and disorder-level aggregation. Deep Sets formalized permutation-invariant functions on unordered sets \citep{zaheer2017}, and attention-based approaches extended this idea to learned set representations and multiple-instance settings \citep{ilse2018,lee2019}. In contrast to these trainable set-aggregation models, our work applies parameter-free infe\-rence-time aggregation to fixed GM-Arc embeddings and distances, avoiding the need to fit an additional aggregation model in a sparse and imbalanced rare-disease setting. The distinction between combining representations before scoring and combining scores after independent analysis has also been studied as early versus late fusion \citep{snoek2005}, which partially motivates our comparison of embedding-level and distance-level patient aggregation.

Aggregating multiple images into a single representation is also common in template-based face recognition, where several images of the same identity in the reference set are pooled before matching \citep{chen2016,hassner2016}. Our setting differs because aggregation is used for rare-disorder prioritization rather than identity recognition, the gallery-side unit is a disorder rather than a person, and many target disorders are represented by only a few patients or are unseen during encoder training. Moreover, these approaches typically aggregate identity-labeled templates in the reference set, whereas our patient-level aggregation combines multiple test images from the same individual before disorder ranking. We therefore focus on inference-time aggregation strategies that operate on the embeddings and distances already produced by the retrieval system.

Disorder-level aggregation is related to prototype- and centroid-based recognition. Prototypical Networks classify samples by comparing them with class prototypes in a learned metric space, providing a simple inductive bias for few-shot recognition \citep{snell2017}. Related few-shot and meta-learning strategies have also been explored for facial phenotype recognition under sparse and imbalanced rare-disorder data \citep{sumer2023}. Nearest-class-mean classifiers similarly represent each class by a mean feature vector and contrast with exemplar-based nearest-neighbor classification \citep{mensink2013}. In our setting, patient-weighted disorder centroids serve a related purpose by summarizing gallery patients for each disorder, but differ in that they are constructed post hoc from fixed facial phenotype embeddings and explicitly account for gallery patients with unequal numbers of images.

At the same time, centroid-based summaries may be insufficient for disorders with variable, age-dependent, or multiple recognizable facial presentations. Exemplar-based nearest-neighbor retrieval retains local image-level information and can capture individual facial presentations that may be lost in a single average representation. This trade-off is particularly relevant in long-tailed recognition, where class-frequency imbalance can affect model behavior \citep{zhang2023}. 
Rare-disorder retrieval represents such a setting, making it important to balance stable disorder-level summaries with local exemplar evidence from individual gallery images. The individual aggregation principles used here are well established in broader machine-learning and face-recognition settings. In rare-disorder facial phenotype retrieval, however, most disorders are represented by only a small number of patients, and many clinically important disorders are absent from encoder training. This makes trainable aggregation models difficult to fit and validate reliably. We therefore evaluate fixed, parameter-free aggregation operators that act at inference time on embeddings and distances already produced by the retrieval system, allowing the framework to be applied without retraining the underlying encoder. To our knowledge, inference-time evidence aggregation has not previously been systematically evaluated for facial phenotype retrieval, and patient-level aggregation, patient-weighted disorder centroids, and hybrid individual-centroid scoring have not been studied jointly in this setting.

\section{Methodology}\label{sec:methods}

We use GM-Arc as a fixed facial phenotype encoder and investigate whether rare-disorder retrieval can be improved through inference-time evidence aggregation. The proposed methods do not modify the model architecture, loss function, or training procedure. Instead, they operate on the embeddings, distances, and disorder-level scores produced by the existing GM-Arc retrieval pipeline.

\subsection{Dataset}\label{sec:dataset}

In this study, we used the GestaltMatcher Database (GMDB) for reproducing the fixed GM-Arc encoder, constructing retrieval galleries, and evaluating aggregation strategies. GMDB v1.1.4 contains 15,381 frontal face portrait images from 11,548 patients across 710 rare genetic disorders. Following the training procedure described in \citet{hustinx2023}, pre-trained ArcFace models were fine-tuned on patient facial images and used to extract facial phenotype embeddings.

Consistent with prior work and the established GMDB evaluation setup, the disorders were divided into two subsets according to the number of available patients. Disorders represented by more than six patients were assigned to GMDB-Freq and included during training. Disorders represented by six or fewer patients were assigned to GMDB-Rare and treated as unseen disorders during evaluation. This distinction allowed us to assess performance both for disorders represented during training and for disorders that were only available through gallery-based retrieval. Throughout, ``frequent'' and ``rare'' refer to how well a disorder is represented in GMDB, rather than to prevalence of rare and ultra-rare disorders.

To specifically evaluate patient-level aggregation, we additionally defined GMDB-Multi, a subset of the GMDB evaluation set containing only patients with more than one available image. This subset contains 996 images from 370 patients across 240 disorders. We further separate this subset into GMDB-Multi-Freq and GMDB-Multi-Rare, depending on whether the corresponding disorder belongs to GMDB-Freq or GMDB-Rare. As a result, each patient in GMDB-Multi-Freq and GMDB-Multi-Rare has on average 3 and 2.5 images, respectively. An overview of the training, gallery, and evaluation splits is shown in Table~\ref{tbl:splits}.

\begin{table}[pos=t]
\caption{Overview of GMDB training, gallery, and evaluation splits. For GMDB-Rare and GMDB-Multi-Rare, values are reported as mean $\pm$ standard deviation across the 10 cross-validation folds where applicable.}\label{tbl:splits}
\footnotesize
\begin{tabular*}{\tblwidth}{@{} LRRR@{} }
\toprule
Dataset & Images & Patients & Disorders \\
\midrule
\multicolumn{4}{@{}l@{}}{\textbf{Training + validation}} \\
GMDB-Freq & 11,128\,+\,1,449 & 8,287\,+\,1,080 & 349 \\
\midrule
\multicolumn{4}{@{}l@{}}{\textbf{Gallery}} \\
GMDB-Freq & 12,577 & 9,367 & 349 \\
GMDB-Rare & 1,098.3\,$\pm$\,7.7 & 879 & 361 \\
GMDB-Freq+Rare & 13,675.3\,$\pm$\,7.7 & 10,246 & 710 \\
\midrule
\multicolumn{4}{@{}l@{}}{\textbf{Evaluation}} \\
GMDB-Freq & 1,255 & 943 & 349 \\
GMDB-Rare & 455.7\,$\pm$\,7.7 & 361 & 361 \\
GMDB-Multi-Freq & 468 & 156 & 105 \\
GMDB-Multi-Rare & 159.9\,$\pm$\,11.8 & 65.2\,$\pm$\,4.4 & 65.2\,$\pm$\,4.4 \\
\bottomrule
\end{tabular*}
\end{table}

During evaluation, the feature space was populated with a fixed gallery of embeddings from diagnosed patients. In the main experiments, we used a unified gallery containing both GMDB-Freq and GMDB-Rare disorders. This setting evaluates each test image against a larger and more heterogeneous reference set and more closely reflects practical diagnostic retrieval, where the relevant disorder subset is not known in advance. In total, GMDB-Rare contained 1,554 images from 1,240 patients across 361 disorders, and was evaluated using 10-fold cross-validation because of the limited number of available patients per disorder. In each fold, an average of 455.7 $\pm$ 7.7 images was used for evaluation and 1,098.3 $\pm$ 7.7 images remained in the corresponding rare-disorder gallery split, which was combined with the full GMDB-Freq gallery (12,577 images) to form the unified gallery of 13,675.3 images on average. The corresponding GMDB-Multi-Rare evaluation subset contained 159.9 $\pm$ 11.8 images from 65.2 $\pm$ 4.4 patients across 65.2 $\pm$ 4.4 disorders per fold, derived from a total of 528 multi-image rare-disorder images from 214 patients across 135 disorders.

The gallery compositions follow the established GMDB evaluation setup and all evaluations were performed such that test patients were excluded from the corresponding gallery fold, preventing identity-level leakage between test and gallery sets.

Because GMDB contains sensitive clinical information and biometric facial images, the underlying patient-level data are not publicly downloadable and are available only through a controlled-access process, as described in the Data and code availability statement.

\subsection{Performance metrics}\label{sec:metrics}

All methods are evaluated as disorder retrieval tasks. For each test image or aggregated test patient, the output is a ranked list of candidate disorders, where a lower rank indicates greater similarity to the test input. For a test image or aggregated test patient, the function $\mathrm{rank}(\cdot)$ denotes the position of the true disorder in this ranked disorder list. Retrieval performance is measured using top-$N$ accuracy for $N \in \{1, 5, 10\}$. A test image or aggregated test patient is considered correctly retrieved at top-$N$ if the true disorder appears within the first $N$ positions of the disorder-level ranking.

Both the number of test images per patient, and patients per disorder differ, as such we balance the metric at both levels. Retrieval outcomes are first averaged within each patient, then averaged within each disorder, and finally averaged over disorders. We refer to this metric throughout as the mean per-disorder top-$N$ accuracy.

Let $\mathcal{D}$ denote the set of disorders represented in the evaluation set, let $\mathcal{P}_d$ denote the set of test patients affected by disorder $d\in\mathcal{D}$, and let $\mathcal{I}_p$ denote the set of test images of patient $p$, with an individual test image written as $t$. For methods that rank each image independently, the patient-level outcome is the fraction of that patient's images for which the true disorder is retrieved at top-$N$:
\begin{equation}\label{eq:image-outcome}
a_{N,p} = 1/|\mathcal{I}_p| \sum_{t\in \mathcal{I}_p} \mathbf{1}[\mathrm{rank}(t)\leq N],
\end{equation}
where $\mathbf{1}[\cdot]$ is the indicator function, equal to 1 if its argument holds and 0 otherwise. This corresponds to the expected top-$N$ accuracy of a single randomly drawn image of patient $p$. For methods that combine multiple images from the same patient before retrieval, such as embedding-level or distance-level patient aggregation, the patient-level outcome is computed from the resulting patient-level disorder ranking: $a_{N,p} = \mathbf{1}[\mathrm{rank}(p) \leq N]$. For patients with a single test image, the independent-image and patient-level definitions coincide.

The per-disorder top-$N$ accuracy is then computed by averaging over patients within each disorder and subsequently over disorders, as:
\begin{equation}\label{eq:per-disorder}
A_{N,d} = 1/|\mathcal{P}_d| \sum_{p\in\mathcal{P}_d} a_{N,p},
\quad
\mathrm{mA}_N = 1/|\mathcal{D}| \sum_{d\in \mathcal{D}} A_{N,d}.
\end{equation}
Here, $\mathrm{mA}_N$ is the mean per-disorder top-$N$ accuracy and $|\mathcal{D}|$ is the number of disorders in the evaluation set.
Averaging within patients before averaging within disorders prevents patients with many test images from dominating the estimate, while averaging across disorders prevents disorders with many test patients from dominating the reported performance.

We additionally report $p$-values computed by comparing the mean per-disorder top-5 accuracy $\mathrm{mA}_5$ of each method with the single-image nearest-neighbor baseline. We used top-5 accuracy for statistical testing because it is less sensitive to rank noise than top-1 accuracy, while remaining more clinically focused than top-10 accuracy. Because methods are evaluated on the same patients, the comparison is paired. The $p$-values are two-sided and obtained from a two-stage cluster bootstrap with $B = 4,000$ resamples, each drawing disorders and then patients within each disorder, mirroring the per-disorder averaging. The same bootstrap resamples were applied to each method. For the rare evaluation set, resampling weights account for patients recurring across the ten cross-validation splits.

\subsection{Baseline single-image retrieval}\label{sec:baseline}

We use GM-Arc as the facial phenotype encoder and the associated retrieval pipeline as the single-image nearest-neighbor baseline. GestaltMatcher disorder prioritization is formulated as retrieval in a facial phenotype embedding space referred to as the Clinical Face Phenotype Space (CFPS). In the baseline setting, each test image is treated independently and compared with a fixed gallery of images from patients with confirmed diagnoses. For each disorder, the retrieval distance is defined by the nearest gallery image belonging to that disorder, and disorders are ranked by increasing distance. We refer to this method as the single-image baseline throughout the remainder of the manuscript. The encoder was reproduced following the published GM-Arc protocol and recommended settings from \citet{hustinx2023}, and updated to GMDB v1.1.4, for which additional implementation details are provided in Supplementary Note~S1.

Given a frontal face photo $x$, the encoder maps the image to an embedding. GM-Arc uses an ensemble of three models and four test-time augmentations (TTA), yielding $R = 12$ embeddings per image, and we write $f_r(x)$ for the embedding of $x$ under the $r$-th model / augmentation combination, $r \in \{1,\ldots,R\}$. Let $G=\{g_1,g_2,\ldots,g_M\}$ denote the set of $M$ gallery images of the evaluation setting under consideration. Each gallery image $g_j$ is associated with a known disorder label $y(g_j)$. For each test-gallery pair and each representation, we compute the cosine distance between the corresponding embeddings:
\begin{equation}\label{eq:cosine}
D_r(t,g_j) = \mathrm{dist}_{\cos}\big(f_r(t),f_r(g_j)\big),
\end{equation}
where $\mathrm{dist}_{\cos}(u,v) = 1 - u^{\top}v \,/\, (\|u\|\,\|v\|) \in [0,2]$ denotes the cosine distance. The gallery images are ranked by increasing cosine distance, where smaller distances correspond to greater embedding similarity and are interpreted as stronger facial phenotypic similarity in CFPS.

Following the original GM-Arc evaluation protocol, these per-representation distances are averaged to obtain the final image-level distance:
\begin{equation}\label{eq:avg-distance}
D(t,g_j) = 1/R \sum_{r=1}^{R} D_r(t,g_j).
\end{equation}
Unless otherwise stated, this average distance is used for all baseline and aggregation experiments. For readability, the definitions that follow are written for a single representation and omit the index $r$. Every embedding-level operation is applied separately within each of the $R$ representations, and the resulting distances are averaged over $r$ as in Eq.~\eqref{eq:avg-distance}.

The image-level gallery ranking is converted into a disorder-level ranking by retaining only the first occurrence of each disorder in the ranked gallery list. For instance, if the nearest gallery images have disorder labels $(A,B,B,C,B)$, the resulting disorder-level ranking is $(A,B,C)$. Thus, the baseline disorder distance for disorder $d$ is equivalent to the distance of the nearest gallery image associated with that disorder:
\begin{equation}\label{eq:nn-distance}
D_{\mathrm{NN}}(t, d) = \min_{g_j\in G_d} D(t, g_j),
\end{equation}
where $G_d\subseteq G$ is the set of gallery images with disorder $d$. Disorders are ranked by increasing $D_{\mathrm{NN}}(t,d)$. This corresponds to $k=1$ nearest-neighbor disorder retrieval.
In the baseline retrieval framework, all test images are evaluated independently. For patients with multiple available test images, each image therefore receives its own disorder-level ranking. As defined in Section~\ref{sec:metrics}, these independent-image retrieval outcomes are averaged within each patient before disorder-level averaging. The aggregation methods introduced in Section~\ref{sec:framework} modify this baseline by changing how gallery embeddings, patient images, or disorder-level scores are aggregated before final disorder ranking.

The limitations of single-image nearest-neighbor retrieval in the CFPS are further characterized in Supplementary Note~S2. There, intra- and inter-disorder cosine distance distributions show that same-disorder image pairs are closer on average than different-disorder pairs, but that the distributions substantially overlap. This overlap motivates aggregation strategies that reduce dependence on isolated nearest-neighbor matches.

\subsection{Proposed multi-level aggregation framework}\label{sec:framework}

Building on the baseline retrieval framework, we define three inference-time aggregation components. First, disorder-level gallery aggregation summarizes multiple diagnosed gallery patients into disorder-level representations. Second, hybrid individual-centroid scoring combines local nearest-neighbor evidence from individual gallery images with global evidence from disorder-level centroids. Third, patient-level aggregation combines multiple test images from the same individual before disorder ranking. These components do not modify the existing GM-Arc encoder.

As illustrated conceptually in Figure~\ref{fig:overview}, the framework contains three aggregation components that operate at different levels of retrieval. For clarity, we first define the gallery-side disorder representations and hybrid scoring function, and then introduce the patient-level aggregation of test images as a replacement for the single-image baseline. The components were chosen to be interpretable, compatible with sparse and imbalanced rare-disorder data, and applicable without training an additional aggregation model.

\subsubsection{Disorder-level gallery aggregation}\label{sec:gallery-aggregation}

\begin{figure}[pos=!ht]
  \centering
  \includegraphics[width=.9\linewidth]{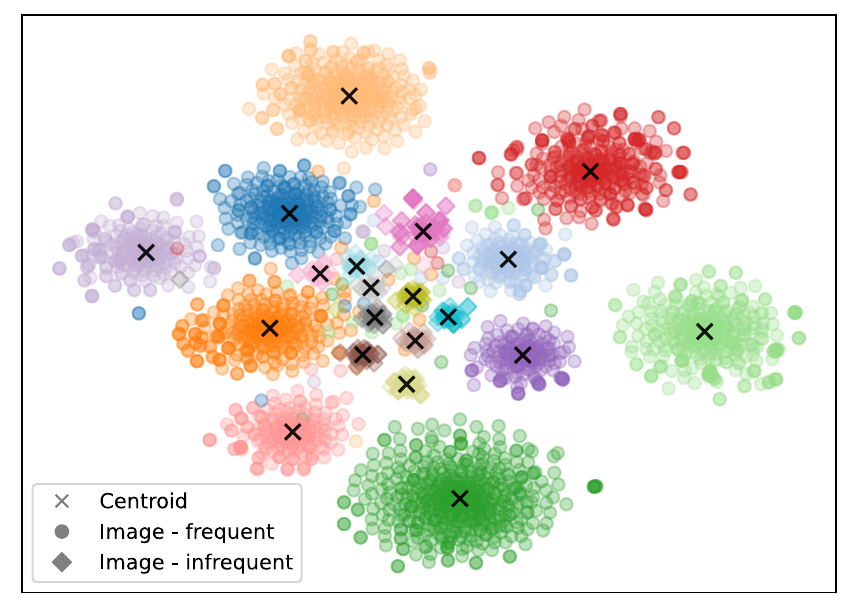}
  \caption{t-SNE visualization of image embeddings and disorder centroids from the 10 most frequent disorders (circle), and 10 less frequent disorders (diamond) in the GMDB-Freq gallery set. Each color represents a disorder, and centroids (X) summarize the corresponding disorder-level embeddings. The central region illustrates that embeddings from different disorders can occupy overlapping regions of the CFPS.}
  \label{fig:tsne}
\end{figure}

The number of gallery patients and images varies substantially across disorders. This imbalance is particularly relevant in the unified-gallery setting, where test images are compared against both GMDB-Freq and GMDB-Rare disorders. When embeddings from different disorders overlap in the CFPS, disorders with many gallery images have more opportunities to contain an individual image that lies close to a test image, even if the disorder as a whole is not the best match. As a result, nearest-neighbor retrieval can be influenced by isolated local matches in overlapping regions of the embedding space rather than by true phenotypic similarity or robust disorder-level evidence. This effect is illustrated qualitatively in Figure~\ref{fig:tsne}, where embeddings from frequent and infrequent disorders show overlapping regions in the CFPS.

To summarize disorder-level gallery evidence, we considered centroid-based representations. These methods do not alter the gallery composition or the underlying embeddings; instead, they change how diagnosed gallery images are represented during retrieval.

\paragraph{Image-weighted disorder centroids}
The image-weighted disorder centroid represents each disorder by the mean embedding of all gallery images associated with that disorder. For a disorder $d$ with gallery image set $G_d$ of size $|G_d|=M_d$, the image-weighted centroid is defined as:
\begin{equation}\label{eq:centroid-img}
\mathrm{centroid}_d^{\,\mathrm{img}} = 1/M_d \sum_{g\in G_d} f(g).
\end{equation}

Retrieval can then be performed by comparing the test-image or test-patient embedding with the centroid of each disorder rather than with individual gallery images.

\paragraph{Patient-weighted disorder centroids}
Because some patients in GMDB have multiple available images, an image-weighted centroid can be biased toward patients with many images. This is particularly relevant for exceptional cases in which a single patient contributes more than ten images, even though most multi-image patients contribute only a small number of images. To avoid this patient-level imbalance, we define a patient-weighted disorder centroid that gives each gallery patient equal weight, regardless of the number of available images.

While this imbalance currently affects only a subset of gallery patients, it is expected to become more relevant as reference databases grow and increasingly contain images from different ages, views, or clinical time points for the same individual. Supplementary Note~S3 summarizes the growth of GMDB across database versions and the increasing representation of patients with multiple images, including highly represented patients with many available images.

For each patient within a disorder, we first compute the mean embedding across that patient's available gallery images. The disorder centroid is then computed as the mean of these patient-level embeddings:
\begin{equation}\label{eq:centroid-pat}
\mathrm{centroid}_d^{\,\mathrm{pat}} = 1/|\mathcal{P}^{\mathrm{gal}}_d| \sum_{p\in\mathcal{P}^{\mathrm{gal}}_d} \left( 1/|G_{d,p}| \sum_{g\in G_{d,p}} f(g) \right),
\end{equation}
where $\mathcal{P}^{\mathrm{gal}}_d$ is the set of gallery patients with disorder $d$, and $G_{d,p} \subseteq G_d$ is the set of gallery images of patient $p$ with that disorder, so that $G_d$ is the union of the $G_{d,p}$ over $p\in\mathcal{P}^{\mathrm{gal}}_d$. Both centroids are therefore determined by the disorder $d$ and by the gallery $G$ in use, which is fixed within a given retrieval run and therefore left implicit in the following notation.

Patient-weighted centroids were used as the primary disorder-level representation because they provide a principled safeguard against over-representing patients with many gallery images. This effect is expected to be modest in the current gallery, where most patients contribute only one or a few images, but may become more relevant as GMDB grows and increasingly contains multiple images per patient. This methodological choice was additionally supported by the held-out GMDB-Freq validation set (1,080 patients), where patient-weighted centroids improved top-1 accuracy by 1.07 to 1.36 percentage points (pp) over image-weighted centroids.

Centroids provide compact summaries of disorder-level gallery evidence, but they are not intended to represent all phenotypic presentations of a disorder completely. For disorders with variable facial expressivity, age-dependent facial manifestations, or multiple recognizable facial presentations, individual gallery images may still provide clinically relevant local exemplar evidence. This motivates the hybrid individual-centroid scoring strategy introduced below.

\subsubsection{Hybrid individual-centroid scoring}\label{sec:hybrid}

Centroid-based retrieval summarizes disorder-level gal\-lery evidence, but a single centroid may not capture disorders with variable facial presentations. Conversely, nearest-neighbor retrieval preserves local exemplar evidence from individual gallery images, but can be sensitive to isolated close matches in overlapping regions of the CFPS. We therefore define hybrid individual-centroid scoring to combine local nearest-neighbor evidence with global disorder-centroid evidence.

For a test image $t$ and disorder $d$, let $D_{\mathrm{NN}}(t,d)$ denote the nearest-neighbor disorder distance defined in Eq.~\eqref{eq:nn-distance}. Let $D_{\mathrm{centroid}}(t,d)$ denote the cosine distance between the test image embedding and the patient-weighted disorder centroid $\mathrm{centroid}_d^{\,\mathrm{pat}}$ defined in Eq.~\eqref{eq:centroid-pat}:
\begin{equation}\label{eq:centroid-distance}
D_{\mathrm{centroid}}(t,d) = \mathrm{dist}_{\cos}\big(f(t), \mathrm{centroid}_d^{\,\mathrm{pat}}\big).
\end{equation}

The hybrid individual-centroid distance is defined as:
\begin{equation}\label{eq:hybrid}
D_{\mathrm{hybrid}}(t,d) = \lambda\, D_{\mathrm{centroid}}(t,d) + (1-\lambda)\, D_{\mathrm{NN}}(t,d),
\end{equation}
where $\lambda \in [0,1]$ controls the relative contribution of global disorder-centroid evidence and local individual-neighbor evidence. With this formulation, $\lambda=0$ corresponds to nearest-neighbor scoring, whereas $\lambda=1$ corresponds to centroid-only scoring. Disorders are ranked by increasing $D_{\mathrm{hybrid}}(t,d)$. When patient-level embedding aggregation is used, $f(t)$ is replaced by the aggregated patient-level embedding defined in Section~\ref{sec:patient-aggregation}.

The hybrid scoring formulation was defined to combine complementary local and global gallery evidence; the weighting parameter $\lambda$ was selected using the held-out GMDB-Freq validation set. We evaluated $\lambda \in \{0.00,\allowbreak 0.25,\allowbreak 0.50,\allowbreak 0.75,\allowbreak 1.00\}$ using a weighted validation objective combining different top-$N$ mean per-disorder accuracy cut-offs, with higher weight assigned to better-rank cut-offs. This procedure selected $\lambda = 0.75$, which was then fixed for all subsequent experiments, including GMDB-Rare and multi-image evaluations. The initial weight-selection procedure is reported in Supplementary Note~S4.1.

A post hoc sensitivity analysis over alternative $\lambda$-values across the evaluation subsets is reported in Supplementary Note~S4.2. Because this analysis used evaluation data, it was not used to select the primary configuration.

\subsubsection{Patient-level aggregation of multiple test images}\label{sec:patient-aggregation}

A single facial image may provide only a partial observation of a patient's phenotype. Embeddings may vary across images of the same individual because of age, head pose, facial expression, illumination, image quality, or clinical time point. In the single-image baseline, these images are evaluated independently. Patient-level aggregation instead combines the available images from the same test patient before disorder ranking, with the aim of reducing dependence on any single image.

Let $\mathcal{I}_p$ denote the set of available test images for patient $p$. For patients with a single test image, patient-level aggregation reduces to the standard single-image retrieval setting. We evaluated two natural fusion points: distance-level fusion, corresponding to late fusion, and embedding-level fusion, corresponding to early fusion.

\paragraph{Distance-level fusion}
Distance-level fusion first scores each test image against every disorder, and then averages these disorder-level distances across the patient's images. Writing $D_{\mathrm{op}}(t,d)$ for the disorder-level distance in use (operator $\mathrm{op}\in \{\mathrm{NN}, \mathrm{centroid}, \mathrm{hybrid}\}$), the patient-level distance is defined as:
\begin{equation}\label{eq:dist-fusion}
D_{\mathrm{dist}}(\mathcal{I}_p,d) = 1/|\mathcal{I}_p| \sum_{t\in\mathcal{I}_p} D_{\mathrm{op}}(t, d).
\end{equation}

Disorders are then ranked by increasing $D_{\mathrm{dist}}(\mathcal{I}_p,d)$. Distance-level fusion is therefore a direct late-fusion extension of the single-image GM-Arc retrieval framework. It preserves the original image-level distance computation and only changes how the resulting disorder-level distances are summarized across multiple test images.

\paragraph{Embedding-level fusion}
As an alternative, we evaluated embedding-level fusion, which aggregates the available test images before retrieval. Instead of computing gallery distances for each image separately, the embeddings of all images from the same test patient are first averaged to form a single patient-level embedding:
\begin{equation}\label{eq:emb-fusion}
\bar{f}(\mathcal{I}_p) = 1/|\mathcal{I}_p| \sum_{t\in\mathcal{I}_p} f(t).
\end{equation}
The patient-level embedding $\bar{f}(\mathcal{I}_p)$ can then be compared with individual gallery images, disorder centroids, or both within the hybrid individual-centroid scoring framework. 

Embedding-level fusion summarizes the patient's available images before retrieval by producing a single patient-level representation in CFPS. This mirrors the patient-level averaging used to construct patient-weighted disorder centroids in the gallery set and provides a consistent representation for nearest-neighbor, centroid-based, and hybrid scoring. Embedding-level fusion was therefore used as the primary patient-level aggregation strategy in the full framework, while distance-level fusion was retained as the natural late-fusion comparator.

\section{Experiments and results}\label{sec:results}

We evaluated the proposed aggregation strategies by first isolating the contribution of each aggregation level and then assessing their combined effect. All main experiments were performed in the unified-gallery setting, where test images were ranked against a combined GMDB-Freq and GMDB-Rare gallery. This setting more closely reflects practical diagnostic retrieval, where the relevant disorder subset is not known in advance.

Because patient-level aggregation is only applicable when multiple test images are available, we first evaluated this component on GMDB-Multi-Freq and GMDB-Multi-Rare. We then evaluated disorder-level gallery aggregation and hybrid individual-centroid scoring on GMDB-Freq and GMDB-Rare, before assessing the full multi-level aggregation framework across all evaluation subsets.

Results are reported as mean per-disorder top-1, top-5, and top-10 accuracy. For GMDB-Rare and GMDB-Multi-Rare, values were averaged across the 10 cross-validation folds. Throughout this section, $p$-values compare mean per-disorder top-5 accuracy between each method and the corresponding GM-Arc single-image nearest-neighbor baseline; values below 0.001 are reported as $p<0.001$.

\subsection{Patient-level aggregation improves multi-image retrieval}\label{sec:res-patient}

We first evaluated whether multiple images from the same test patient improve disorder retrieval. This analysis was performed on GMDB-Multi-Freq and GMDB-Multi-Rare, which contain only test patients with more than one available image. We compared the single-image baseline with distance-level and embedding-level patient aggregation. This baseline was summarized at the patient level by first averaging retrieval outcomes across images from the same patient, then averaging within and across disorders. The baseline therefore estimated the accuracy expected from one randomly drawn image of the patient, and is directly comparable to the patient-level aggregation approaches.

As shown in Table~\ref{tbl:patient-aggregation}, both fusion strategies improved retrieval compared with the single-image baseline on GMDB-Multi-Freq. Distance-level and embedding-level fusion performed similarly at top-1, with 55.95\% and 56.04\% accuracy, respectively, compared with 46.12\% for the baseline. Distance-level fusion achieved higher top-5 and top-10 accuracy on this subset. However, on GMDB-Multi-Rare, distance-level fusion decreased top-1 accuracy from 18.54\% to 17.75\%, whereas embedding-level fusion preserved top-1 accuracy and improved top-5 and top-10 accuracy compared with both the baseline and distance-level fusion.

\begin{table*}[width=0.9\FullWidth,pos=t]
\caption{Patient-level aggregation results on GMDB-Multi-Freq and GMDB-Multi-Rare using the unified gallery. Results are reported as mean per-disorder top-$N$ accuracy. Reported $p$-values compare mean per-disorder top-5 accuracy between each aggregation method and the corresponding single-image baseline.}\label{tbl:patient-aggregation}
\begin{tabular*}{\tblwidth}{@{} LLRRRR@{} }
\toprule
Approach & Evaluation set & Top-1 & Top-5 & Top-10 & $p$-value \\
\midrule
Single-image baseline & GMDB-Multi-Freq & 46.12 & 66.61 & 74.21 & n/a \\
Distance-level fusion & & 55.95 & \textbf{77.94} & \textbf{80.63} & $<$0.001 \\
Embedding-level fusion & & \textbf{56.04} & 76.29 & 77.97 & $<$0.001 \\
\midrule
Single-image baseline & GMDB-Multi-Rare & \textbf{18.54} & 27.62 & 32.58 & n/a \\
Distance-level fusion & & 17.75 & 30.43 & 36.24 & 0.056 \\
Embedding-level fusion & & 18.53 & \textbf{32.10} & \textbf{36.33} & 0.004 \\
\bottomrule
\end{tabular*}
\end{table*}

These results empirically support the use of embedding-level fusion as the patient-level aggregation strategy in the full framework. Although distance-level fusion performed strongly on GMDB-Multi-Freq, embedding-level fusion provided more stable behavior across both GMDB-Multi-Freq and GMDB-Multi-Rare subsets, consistent with its role as a single patient-level CFPS representation. Additional stratified analyses in patients with increasing numbers of available images showed that the benefit of patient-level aggregation increased with the number of images used on GMDB-Multi-Freq, with the largest gain from adding a second image and diminishing returns thereafter. On GMDB-Multi-Rare, the corresponding gains were smaller, indicating that the benefit of additional test images was less pronounced on GMDB-Multi-Rare. Detailed results are provided in Supplementary Note~S5.

To further contextualize patient-level aggregation, we compared the fusion approaches with a retrospective best-single-image reference. For each multi-image test patient, this reference selected the individual image with the best rank of the true disorder. Because this selection used ground-truth information, it served as an optimistic single-image selection reference. On average, the best-single-image reference outperformed both fusion strategies. However, both distance-level and embedding-level fusion achieved a better true-disorder rank than this reference for a subset of patients. Detailed results are provided in Supplementary Note~S6.

\subsection{Disorder-level gallery aggregation improves retrieval}\label{sec:res-gallery}

We next evaluated whether aggregating gallery images at the disorder level improves retrieval. This analysis was performed on GMDB-Freq and GMDB-Rare to assess whether centroid-based gallery representations improve performance for both disorders represented during training and unseen rare disorders. We compared the single-image nearest-neighbor baseline with two disorder-level aggregation strategies: image-weighted disorder centroids and patient-weighted disorder centroids.

As shown in Table~\ref{tbl:gallery-aggregation}, both centroid-based approaches improved performance compared with the single-image baseline. On GMDB-Freq, baseline top-1 accuracy was 38.52\%, whereas image-weighted and patient-weighted disorder centroids achieved 45.77\% and 46.34\%, respectively. On GMDB-Rare, top-1 accuracy increased from 19.38\% with baseline retrieval to 22.13\% with image-weighted centroids and 22.88\% with patient-weighted centroids. Stratification by gallery disorder size showed that these centroid-based gains were concentrated among sparsely represented disorders, suggesting that disorder-level summaries are particularly useful when only few gallery images are available for a disorder (Supplementary Note~S7).

\begin{table*}[width=0.9\FullWidth,pos=t]
\caption{Disorder-level gallery aggregation and hybrid individual-centroid scoring results on GMDB-Freq and GMDB-Rare using the unified gallery. Results are reported as mean per-disorder top-$N$ accuracy; $p$-values compare mean per-disorder top-5 accuracy between each aggregation method and the corresponding single-image baseline.}\label{tbl:gallery-aggregation}
\begin{tabular*}{\tblwidth}{@{} LLRRRR@{} }
\toprule
Approach & Evaluation set & Top-1 & Top-5 & Top-10 & $p$-value \\
\midrule
Single-image baseline & GMDB-Freq & 38.52 & 58.35 & 65.47 & n/a \\
Image-weighted disorder centroid & & 45.77 & 64.80 & 73.80 & $<0.001$ \\
Patient-weighted disorder centroid & & 46.34 & 65.26 & 73.34 & $<0.001$ \\
Hybrid individual-centroid scoring & & \textbf{47.03} & \textbf{66.86} & \textbf{74.90} & $<0.001$ \\
\midrule
Single-image baseline & GMDB-Rare & 19.38 & 30.18 & 34.90 & n/a \\
Image-weighted disorder centroid & & 22.13 & 38.02 & 44.86 & $<0.001$ \\
Patient-weighted disorder centroid & & 22.88 & \textbf{38.39} & \textbf{45.33} & $<0.001$ \\
Hybrid individual-centroid scoring & & \textbf{23.10} & 38.08 & 44.40 & $<0.001$ \\
\bottomrule
\end{tabular*}
\end{table*}

Patient-weighted disorder centroids achieved the highest top-1 performance among the centroid-based variants on both evaluation subsets, although the improvement over image-weighted centroids was modest. This modest difference is consistent with the current gallery composition, where most patients contribute only one or a few images. Hybrid individual-centroid scoring results are shown in the same table and discussed in Section~\ref{sec:res-hybrid}.

\subsection{Hybrid individual-centroid scoring improves disorder ranking}\label{sec:res-hybrid}

We next evaluated the hybrid individual-centroid scoring strategy on the main evaluation subsets. In this experiment, the single-image baseline corresponds to GM-Arc nearest-neighbor scoring, where each disorder is represented by its closest individual gallery image. We compared this baseline with centroid-only scoring and hybrid individual-centroid scoring using the patient-weighted disorder centroids defined in Section~\ref{sec:gallery-aggregation}. Because the hybrid weighting parameter $\lambda = 0.75$ was selected using only the held-out GMDB-Freq validation set, this analysis assessed whether the same local-global scoring strategy generalizes beyond validation and transfers to unseen rare disorders. 

As shown in Table~\ref{tbl:gallery-aggregation}, hybrid individual-centroid scoring improved retrieval compared with the single-image baseline on both evaluation subsets. On GMDB-Freq, top-1 accuracy increased from 38.52\% with the single-image baseline and 46.34\% with centroid-only scoring to 47.03\% with hybrid scoring. Similar numerical gains were observed for top-5 and top-10 accuracy. On GMDB-Rare, hybrid scoring achieved the highest top-1 accuracy, increasing from 19.38\% with the single-image baseline and 22.88\% with centroid-only scoring to 23.10\%. However, centroid-only scoring achieved slightly higher top-5 and top-10 accuracy than hybrid scoring on GMDB-Rare.

These results indicate that hybrid scoring transfers well to GMDB-Freq test disorders, consistent with the validation-set behavior, but provides more limited additional benefit for unseen rare disorders. This pattern is consistent with the gallery-size stratification analysis, where the gain of hybrid scoring over centroid-only scoring was concentrated among well represented disorders, whereas GMDB-Rare disorders remained largely in the sparsely represented regime (Supplementary Note~S7). The post hoc sensitivity analysis over $\lambda$-values further supports this interpretation, showing that more centroid-dominant weighting can be preferable for some rare-disorder settings (Supplementary Note~S4.2).

\subsection{Full multi-level aggregation improves retrieval across evaluation subsets}\label{sec:res-full}

\begin{figure*}[pos=!ht]
  \centering
  \includegraphics[width=\linewidth]{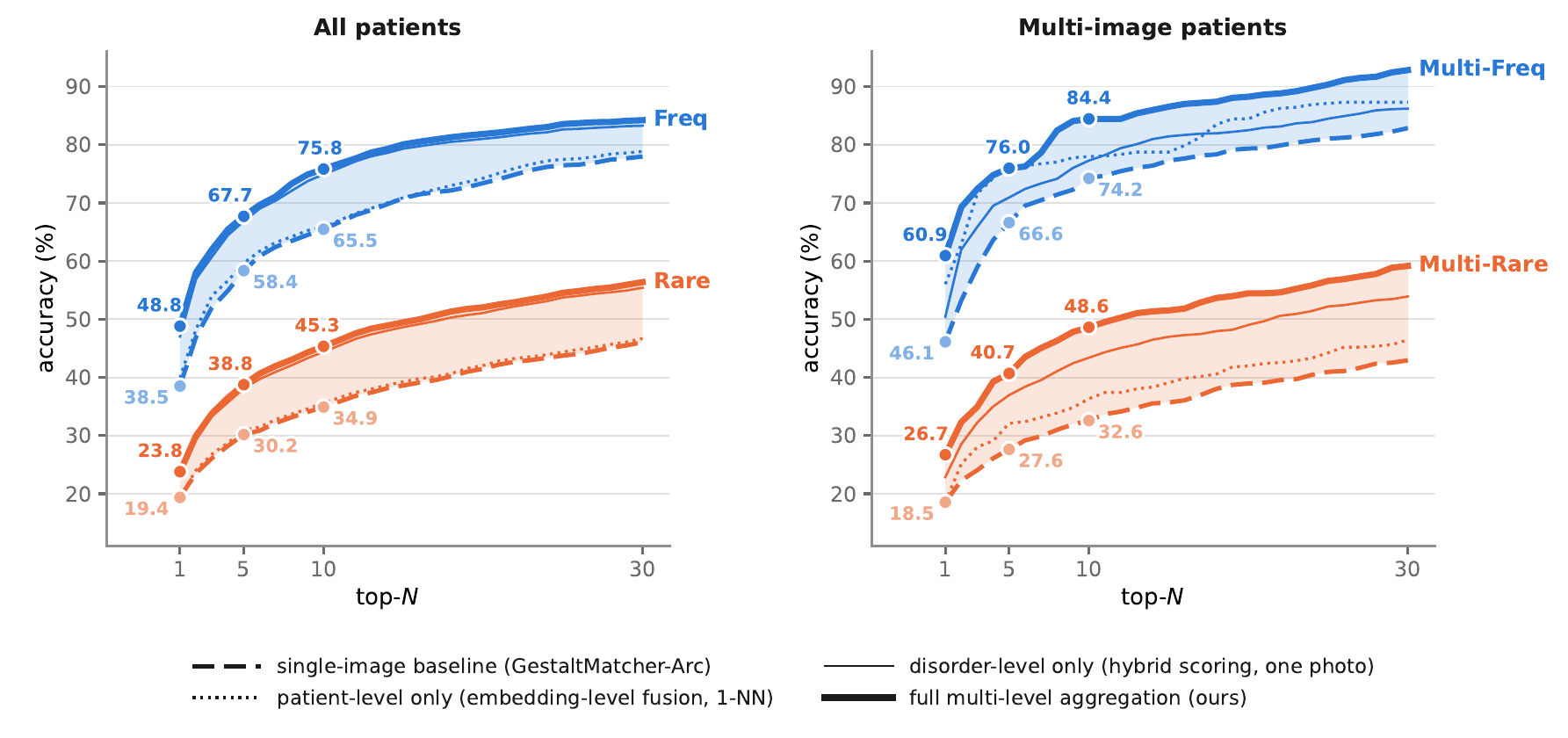}
  \caption{Mean per-disorder top-$N$ accuracy of the single-image nearest-neighbor baseline and of full multi-level aggregation, on the unified gallery. Left panel: all-patient evaluation sets GMDB-Freq and GMDB-Rare. Right panel: Multi-image subsets GMDB-Multi-Freq and GMDB-Multi-Rare. Dashed lines are the baseline, dotted lines the patient-level aggregation only, thin lines the disorder-level aggregation only, and solid lines the full aggregation. Dots mark top-1, top-5 and top-10 on the baseline and the full-aggregation curve, labeled with the corresponding accuracies in percent.}
  \label{fig:headline}
\end{figure*}

Finally, we evaluated the full multi-level aggregation framework across all evaluation subsets. The full framework combines the primary components defined in Section~\ref{sec:framework}: patient-weighted disorder centroids, hybrid individual-cen\-troid scoring with $\lambda = 0.75$, and embedding-level patient aggregation.

As shown in Figure~\ref{fig:headline}, full multi-level aggregation improved retrieval performance across all evaluation subsets. On GMDB-Freq, top-1 accuracy increased from 38.52\% with single-image baseline retrieval to 48.82\% with full aggregation (+10.30 pp). On GMDB-Rare, top-1 accuracy increased from 19.38\% to 23.79\% (+4.41 pp). On the multi-image subsets, top-1 accuracy increased from 46.12\% to 60.94\% on GMDB-Multi-Freq (+14.82 pp) and from 18.54\% to 26.71\% on GMDB-Multi-Rare (+8.17 pp). Overall, full aggregation significantly outperformed the corresponding single-image baseline on all evaluation subsets (GMDB-Freq: $p < 0.001$; GMDB-Rare: $p < 0.001$; GMDB-Multi-Freq: $p = 0.003$; GMDB-Multi-Rare: $p < 0.001$).  

\begin{figure*}[pos=!ht]
  \centering
  \includegraphics[width=\linewidth]{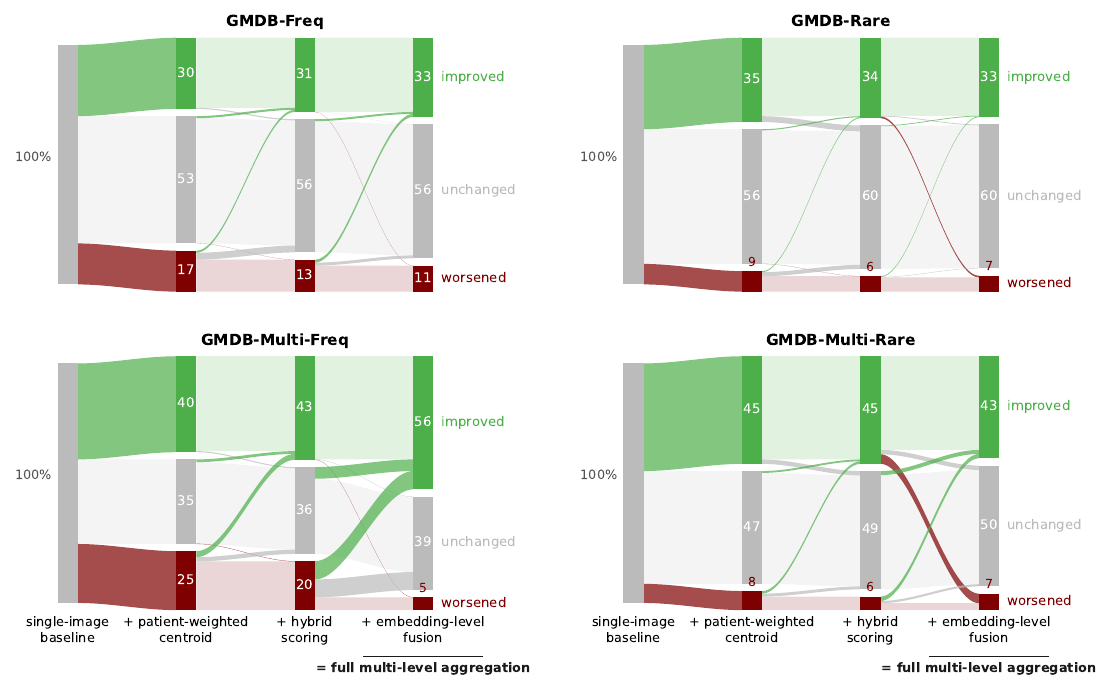}
  \caption{Per-patient change in the rank of the correct disorder relative to the single-image nearest-neighbor baseline rank, on the unified gallery. Numbers give the percentage of test patients whose rank improves (green), worsens (red), or is unchanged (gray). Flows follow the same patients as one aggregation component is added at a time, from the single-image baseline on the left to the full multi-level aggregation on the right. In the GMDB-Freq and GMDB-Rare evaluation sets, the large unchanged fraction of the embedding-level fusion is dominated by single-image patients, for whom patient-level fusion cannot alter the score. Ranks are censored at 30.}
  \label{fig:rank-change}
\end{figure*}

To assess whether the aggregate performance gains reflected broadly distributed improvements rather than large gains in a small number of cases, we compared patient-level true-disorder ranks between the baseline and the selected aggregation approaches. For each test patient, we recorded whether the rank of the correct disorder improved, worsened, or remained unchanged relative to the nearest-neighbor baseline. Because ranks worse than 30 are often not considered clinically relevant, ranks were censored at 30 before averaging, and rank changes outside that cut-off were ignored. As shown in Figure~\ref{fig:rank-change}, improvements outweighed degradations at every stage and in all evaluation subsets. For the full multi-level aggregation framework, the true-disorder rank improved for 55.6\% of patients on GMDB-Multi-Freq and 42.6\% of patients on GMDB-Multi-Rare, while worsening for only 5.4\% and 6.9\%, respectively. In the full GMDB-Freq and GMDB-Rare evaluation sets, the corresponding proportions were 33.1\% improved versus 10.9\% worsened, and 33.2\% improved versus 6.7\% worsened, respectively. Disorder-level gallery aggregation alone already improved the rank for 29.7\% to 44.9\% of patients across the four subsets, and adding hybrid scoring and patient-level aggregation reduced the worsened fraction in every subset, most markedly on GMDB-Multi-Freq, where it fell from 24.6\% to 5.4\% while the improved fraction rose from 40.1\% to 55.6\%. Repeating the same comparison on uncensored ranks raised the improved fraction to 70.5\% on GMDB-Rare and 81.1\% on GMDB-Multi-Rare (against 11.4\% and 5.7\% worsened), and lowered the median true-disorder rank from 49 to 17 and from 72 to 12, respectively.

This indicates that the observed accuracy gains reflect a broad shift toward improved patient-level rankings rather than isolated improvements in a small number of cases.

The contribution of individual aggregation components varied across evaluation subsets, cut-offs, and gallery settings. In general, the full framework performed best or among the best configurations, but the component effects were not consistently additive. This suggests that patient-level aggregation, disorder-level gallery aggregation, and hybrid individual-centroid scoring capture partly overlapping evidence, while still providing complementary benefits in the fixed full configuration. The full framework was evaluated as a single pre-defined configuration rather than optimized separately for each evaluation subset. Detailed component-wise ablations in the unified-gallery and split-gallery settings are provided in Supplementary Note~S8.

\section{Discussion}\label{sec:discussion}

This study evaluated whether facial phenotype retrieval for rare genetic disorder prioritization can be improved through inference-time evidence aggregation, without modifying the underlying GM-Arc encoder. The proposed framework aggregates evidence across multiple test images, dis\-order-level gallery representations, and hybrid individual-centroid scores. We discuss the main findings across GMDB evaluation subsets, their relevance for AI-assisted rare-disorder prioritization, and the limitations that should guide future work.

\subsection{Main findings and interpretation}\label{sec:disc-findings}

The component analyses show that inference-time aggregation affects retrieval differently depending on whether evidence is aggregated across test images, disorder-level gallery representations, or disorder-level scores. These differences are important for interpreting where the observed performance gains arise and why their magnitude varies across GMDB evaluation subsets.

The patient-level results show that multiple images of the same individual can provide useful retrieval evidence, but that the fusion strategy matters. Distance-level fusion performed strongly on GMDB-Multi-Freq, whereas embedding-level fusion showed more stable behavior across both frequent and unseen rare multi-image subsets. This supports the use of a single patient-level CFPS representation in the full framework and is consistent with the idea that forming such a representation before retrieval can reduce dependence on individual images and shift the patient's position in CFPS in a way that is not captured by averaging image-level distances alone. The retrospective best-single-image reference described in Supplementary Note~S6 further shows that patient-level aggregation is not simply equivalent to selecting the best available image. Although this reference performed better on average, fusion improved the true-disorder rank for a subset of patients, indicating that aggregation can alter the full disorder ranking rather than only recover the strongest individual image.

The disorder-level results show that centroid-based gal\-lery representations consistently improved retrieval over nearest-neighbor scoring. This supports the idea that summarizing disorder-level evidence can reduce dependence on isolated local matches in overlapping regions of CFPS. The gallery-size stratification analysis further showed that centroid-based gains were concentrated among sparsely represented disorders, supporting the use of disorder-level summaries when individual gallery evidence is limited (Supplementary Note~S7). The modest difference between image-weighted and patient-weighted centroids is consistent with our expectation for the current gallery composition, where most patients contribute only one or a few images. Patient-weighting therefore acts mainly as a principled safeguard against patient-level imbalance, rather than as a source of large immediate performance gains, and is expected to become more relevant as GMDB grows.

The hybrid-scoring results indicate that the balance between local exemplar evidence and disorder-level summaries depends on gallery representation. The validation-selected hybrid weighting transferred well to GMDB-Freq, where hybrid scoring improved over both nearest-neighbor and centroid-only scoring. For GMDB-Rare, however, centroid-only scoring already captured most of the gain over nearest-neighbor retrieval, and hybrid scoring provided less consistent additional benefit at higher rank cut-offs. This pattern supports the view that local nearest-neighbor evidence is most useful when disorders are sufficiently represented in the gallery, whereas unseen rare disorders remain closer to a sparse-gallery regime in which centroid-based scoring captures most of the available disorder-level signal.

The unified-gallery setting further highlights that retrieval performance depends on the composition of the reference gallery. Because GMDB-Freq and GMDB-Rare disorders are ranked together, aggregation strategies can affect well represented and sparsely represented disorders differently. Hybrid scoring benefits more from local nearest-neighbor evidence when disorders are sufficiently represented, whereas rare-disorder retrieval appears to depend more strongly on disorder-level summaries. This creates a trade-off in which a single fixed scoring strategy must balance performance across gallery regions with different levels of representation.

The full multi-level framework improved retrieval across all evaluation subsets, with the largest gains observed for multi-image patients. However, the component effects were not uniformly additive. Patient-level and disorder-level aggregation can provide combined gains, but they also partly overlap in the evidence they use, and their relative contributions shift across evaluation settings. For instance, on GMDB-Multi-Rare, embedding-level patient aggregation had little isolated top-1 effect, but contributed more clearly when combined with disorder-level aggregation and hybrid scoring. One interpretation is that, for unseen disorders in GMDB-Rare, aggregating the test patient alone is insufficient when the corresponding gallery representation remains sparse and unstable. Once gallery-side evidence is summarized through disorder-level centroids and combined with local exemplar evidence, a single patient-level CFPS representation may become more useful for ranking the correct disorder. More broadly, these findings support viewing patients and disorders as regions sampled by multiple observations in CFPS, rather than as single fixed points. Patient-level aggregation combines multiple observations of the same individual, while disorder-level aggregation summarizes multiple diagnosed observations of the same disorder.

\subsection{Clinical relevance}\label{sec:disc-clinical}
The proposed framework is intended to improve disorder prioritization through retrieval, rather than to provide an autonomous diagnosis. In AI-assisted facial phenotyping, ranked candidate disorders can support clinicians by prioritizing conditions for further review, targeted follow-up genetic testing, or phenotype-driven variant interpretation. Improving retrieval robustness is therefore clinically relevant for rare disorder evaluation, especially when a disorder is represented in the reference gallery by only a small number of patients with confirmed diagnoses, or when individual test or gallery images are not fully representative. Although the absolute gains are most directly interpretable, they also correspond to substantial relative increases in mean per-disorder top-1 accuracy. For example, the 14.82 pp top-1 gain on GMDB-Multi-Freq corresponds to a 32.1\% relative increase in mean per-disorder top-1 accuracy.

The results suggest that retrieval can benefit from aggregating facial phenotype evidence across the levels at which such evidence is available in practice. Patients may have multiple photographs acquired at different ages, with different head poses, under different imaging conditions, or from different viewpoints. Standardized image acquisition may also be difficult in some clinical contexts, particularly in pediatric or neurodevelopmental rare-disease cohorts. Similarly, disorders are represented by collections of diagnosed cases rather than by a single canonical facial presentation. Patient-level aggregation and disorder-level gallery aggregation therefore align more closely with how facial phenotype evidence accumulates in clinical and reference-database settings: across multiple observations of an individual and across multiple diagnosed reference cases for a disorder.

The unified-gallery results are particularly relevant in this context. In practice, whether the correct disorder is well represented or sparsely represented in the reference database is not known before ranking, and a test patient must be compared against a broad and imbalanced set of candidate disorders. The improvements observed in the unified gallery suggest that multi-level aggregation may be useful when rare-disorder cases are evaluated against larger reference sets containing disorders with different levels of gallery representation. However, the clinical impact of these improvements should be assessed in prospective studies that measure effects on diagnostic workflows, candidate-gene prioritization, variant interpretation, and clinical decision-making.

\subsection{Limitations and future work}\label{sec:disc-limitations}

Several limitations should be considered. The number of evaluation patients, particularly multi-image evaluation patients, remains limited. Although the results indicate that patient-level aggregation can improve retrieval when multiple images of the same individual are available, larger evaluation sets are needed to estimate these effects more precisely and to determine how performance depends on the number, age range, clinical time span, and quality of available images per patient. Initial exploratory analyses of factors such as image number and patient age range did not reveal strong correlations with rank improvement, but these analyses were preliminary and likely underpowered (Supplementary Note~S9).

The proposed methods also remain constrained by the fixed GM-Arc embedding space. This design isolates the effect of inference-time aggregation, but it also means that performance depends on the structure of the learned CFPS. If certain disorders, age groups, ancestries, or facial presentation subgroups are poorly separated in the embedding space, aggregation alone may not fully resolve these limitations. For instance, overlap between ground-truth and non-ground-truth distance distributions illustrates an intrinsic limitation of single-image nearest-neighbor retrieval (Supplementary Note~S2). Aggregation changes the decision statistic by combining evidence across images, gallery representations, or disorder-level scores, but it cannot fully compensate for an embedding space in which clinically distinct disorders are poorly separated.

Next, the disorder-level aggregation methods use simple arithmetic averages in embedding space. Patient-weighted centroids provide a compact and interpretable disorder-level representation, but a single centroid may be insufficient for disorders with variable facial expressivity, age-dependent facial manifestations, subtype-specific facial patterns, or multiple recognizable facial presentations. Hybrid individual-centroid scoring partly addresses this limitation by retaining local nearest-neighbor evidence, but it does not explicitly model disorder substructure. Related exploratory experiments with proportional nearest-neighbor scoring, which aimed to reduce sensitivity to isolated gallery matches by averaging over a disorder-specific fraction of nearest neighbors, are reported in Supplementary Note~S10. Future work should investigate subtype-aware, multi-centroid, or density-based gallery representations as larger reference datasets become available.

The full framework uses fixed aggregation choices rather than adaptive weighting. The hybrid weighting parameter was selected on the held-out GMDB-Freq validation set and then fixed for all subsequent evaluations. This avoids selecting the primary configuration on the evaluation subsets, but the validation set contains only disorders represented during training and does not fully capture the unseen-disorder setting. The results suggest that the optimal balance between nearest-neighbor and centroid-based evidence may depend on gallery representation, with well represented disorders benefiting more from local nearest-neighbor evidence and sparsely represented disorders relying more strongly on centroid-based summaries. Future work could therefore investigate gallery-size-adaptive or disorder-dependent weighting strategies, as well as image-level weighting for patients with multiple images and model- or test-time-augmentation weighting. Such strategies would need to be selected and validated independently to avoid overfitting to the current GMDB evaluation subsets.

A further limitation is that the evaluation assumes a single ground-truth disorder label per test patient. In clinical genetics, some patients may have dual or multiple molecular diagnoses, resulting in blended or composite phenotypes \citep{posey2017}. Such cases may not be well represented by a retrieval framework that ranks candidate disorders against a single reference diagnosis, and standard top-N accuracy may not capture whether the system retrieves one component of a dual diagnosis. Future evaluations should therefore consider multi-label or composite-diagnosis cases when sufficiently curated data become available.

Finally, the evaluation was based on a single database and model ensemble. All experiments used GMDB v1.1.4 and a reproduced GM-Arc ensemble, and the proposed operators were evaluated within this specific GMDB/GM-Arc retrieval setting. This limits the extent to which the results can be assumed to generalize to other databases, alternative facial phenotype encoders, or prospective clinical workflows. At the same time, this limitation reflects the current state of the field: to our knowledge, GMDB is the only large, structured, and accessible rare-disorder facial phenotype database suitable for this type of systematic retrieval evaluation. Although the split-gallery analyses show that the observed gains persist under a substantial change in gallery composition (Supplementary Note~S8), this does not constitute external validation. External cohorts, future GMDB releases, and alternative encoders will therefore be important for assessing whether the same aggregation principles and weighting choices generalize beyond the present benchmark.

\section{Conclusion}\label{sec:conclusion}

This study shows that facial phenotype retrieval for rare genetic disorder prioritization can be improved through inference-time evidence aggregation without modifying the underlying GestaltMatcher-Arc encoder. The proposed framework aggregates evidence across multiple test images, dis\-order-level gallery representations, and hybrid individual-centroid scores.

Across GMDB evaluation subsets, the framework improved mean per-disorder retrieval accuracy in the unified-gallery setting, where test cases were ranked against a heterogeneous reference gallery containing both GMDB-Freq and GMDB-Rare disorders. Because the proposed aggregation strategies operate at inference time and require no retraining, they can be applied to existing facial phenotype encoders already in use. These findings support a shift from isolated single-image matching toward multi-level aggregation of patient and disorder evidence in facial phenotype retrieval.

Given that the evaluation was retrospective and based on GMDB v1.1.4 with a reproduced GM-Arc encoder, future work should assess whether these aggregation principles generalize to future database versions, alternative encoders, and prospective clinical rare-disease workflows.


\section*{Data and code availability}

The GMDB v1.1.4 data used in this study are available to qualified researchers for the development and evaluation of next-generation phenotyping methods through a controlled-access process. Researchers seeking access must obtain appropriate ethics approval, submit a research proposal, and sign a data-use agreement that ensures compliance with applicable data protection legislation, including the General Data Protection Regulation (GDPR). Requests are evaluated by the GMDB Advisory Board and are subject to applicable legal, ethical, consent, and institutional requirements. Because GMDB contains sensitive clinical information and facial images that may permit identification of individuals with rare genetic disorders, the underlying patient-level data cannot be made publicly available.

Code for reproducing the GM-Arc model ensemble, inference-time aggregation, and evaluation procedures will be made available upon journal acceptance, excluding restricted patient data and any model or database components that cannot be redistributed. Researchers with approved GMDB access can use the reported protocol and code to reproduce the analyses on the corresponding controlled-access data.

\printcredits

\section*{Declaration of competing interest}

The authors declare that they have no known competing financial interests or personal relationships that could have appeared to influence the work reported in this paper.

\section*{Funding}
BJ is funded by the German Federal Ministry of Research, Technology, and Space (Bone2Gene project, GO-Bio initial Machbarkeitsphase 4).

\section*{Declaration of generative AI and AI-assisted technologies in the manuscript preparation process}

During the preparation of this work, the authors used OpenAI ChatGPT to support language editing, restructuring, and refinement of text. The tool was not used to generate results, perform analyses, create or modify datasets, or make scientific decisions. All content, references, and claims were checked, verified by the authors against the original text, and edited as needed. We take full responsibility for the content of the publication.

Figure~\ref{fig:overview} includes a synthetic facial image generated using GestaltGAN \citep{kirchhoff2025} for explanatory purposes only. The image does not depict a real patient, was not used as research data, and was reviewed by the authors for appropriateness and accuracy of the representation.

\putbib
\end{bibunit}

\clearpage
\onecolumn
\ResetForAppendix

\renewcommand{\thesection}{S\arabic{section}}
\renewcommand{\thesubsection}{S\arabic{section}.\arabic{subsection}}
\renewcommand{\thefigure}{S\arabic{figure}}
\renewcommand{\thetable}{S\arabic{table}}
\renewcommand{\theequation}{S\arabic{equation}}

\makeatletter
\setlength{\@fptop}{0pt}
\setlength{\@fpsep}{8pt plus 1fil}
\setlength{\@fpbot}{0pt plus 1fil}
\setlength{\@dblfptop}{0pt}
\setlength{\@dblfpsep}{8pt plus 1fil}
\setlength{\@dblfpbot}{0pt plus 1fil}
\makeatother

\RenewDocumentEnvironment{Abstract}{o}{\setbox0=\vbox\bgroup}{\egroup}
\RenewDocumentCommand{\dashrule}{O{.4pt}mm}{}

\begin{bibunit}
\csdef{lastpage}{\pageref{LastPage}}
\let\WriteBookmarks\relax
\def\floatpagepagefraction{1}
\def\textpagefraction{.001}

\shorttitle{Appendix A: Supplementary notes}
\shortauthors{A. Hustinx, C. Kaffin\'e et~al.}

\title[mode = title]{Appendix A: Supplementary notes for \emph{Multi-Level Evidence Aggregation for Robust Facial Phenotype Retrieval in Rare Genetic Disorder Prioritization}}

\author[1]{Alexander Hustinx}[orcid=0000-0003-4592-3979]
\fnmark[1]
\cormark[1]
\ead{ahustinx@uni-bonn.de}

\author[1]{Carolin Kaffin\'e}[orcid=0009-0003-3896-6847]
\fnmark[1]

\author[1]{Behnam Javanmardi}[orcid=0000-0002-9317-6114]

\author[1]{Tzung-Chien Hsieh}[orcid=0000-0003-3828-4419]

\author[1]{Peter Krawitz}[orcid=0000-0002-3194-8625]

\affiliation[1]{organization={Institute for Genomic Statistics and Bioinformatics, University Hospital Bonn},
                city={Bonn},
                country={Germany}}

\fntext[1]{Authors contributed equally.}
\cortext[1]{Corresponding author.}

\thispagestyle{first}
\MaketitleBox
\printFirstPageNotes
\vspace{1cm}
\section*{Supplementary note overview}
Throughout this work, the main experiments were conducted on the GestaltMatcher Database (GMDB), a controlled-access dataset consisting of frontal facial images of individuals affected by rare and ultra-rare genetic disorders \citep{lesmann2024}. GMDB was introduced by \citet{hsieh2022} to support the development and evaluation of GestaltMatcher (GM), a rare-disorder retrieval framework that compares individual test images with gallery images from patients with confirmed diagnoses. GestaltMatcher formulates facial phenotype analysis as retrieval in the Clinical Face Phenotype Space (CFPS), where visually and phenotypically similar cases are expected to be located closer together \citep{hsieh2022}. 

\citet{hustinx2023} subsequently updated the model architecture and training strategy to improve retrieval and verification performance for disorders represented during training (GMDB-Freq) and unseen disorders (GMDB-Rare). This implementation is referred to as GestaltMatcher-Arc (GM-Arc), reflecting its use of ArcFace-based facial representation models \citep{deng2019}. In our work, reproduced GM-Arc models served as the fixed facial phenotype encoder and as the basis for the single-image baseline.

This supplementary document provides implementation details, additional methodological analyses, sensitivity analyses, ablation experiments, and secondary evaluations supporting the main manuscript. Specifically, the supplementary notes describe the reproduced GM-Arc implementation, the increasing number of multi-image patients in GMDB, hybrid-scoring weight selection and sensitivity, component-wise ablations, exploratory analyses of patient-level aggregation, intra- and inter-disorder distance distributions, stratified multi-image performance, centroid performance by gallery representation size, comparison with a retrospective best-single-image reference, and proportional nearest-neighbor scoring.

\clearpage
\section{Baseline GestaltMatcher-Arc implementation details}\label{supp:sec:implementation}

The GestaltMatcher-Arc (GM-Arc) ensemble described by \citet{hustinx2023} consists of three models: two models fine-tuned on GMDB, denoted m0 and m1, and one iResNet-100 ArcFace model trained on GLINT360K, denoted m2 \citep{an2021,deng2019}. In this work, m0 and m1 were retrained using GMDB v1.1.4, while m2 was used as the external ArcFace model. Faces were detected and aligned to the ArcFace five-landmark reference with RetinaFace before encoding \citep{deng2020}. For each image, the three-model ensemble was combined with four test-time augmentations, yielding 12 model/TTA-specific embeddings per image. As a base for reproducing the models we used their official code repository: \url{https://github.com/igsb/GestaltMatcher-Arc/tree/wacv2023}.

Supplementary Table~\ref{supp:tbl:implementation} summarizes the implementation details used to reproduce the baseline GM-Arc encoder and retrieval framework. These settings include the model ensemble, training configuration, inference protocol, distance metric, and baseline disorder-scoring procedure. All hyperparameters originally selected by \citet{hustinx2023} were used for retraining GM-Arc. Unless otherwise stated, the same fixed encoder and inference settings were used for all aggregation experiments in the main manuscript.

\begin{table}[width=\linewidth,cols=2,pos=!ht]
\caption{Implementation details for the reproduced GM-Arc encoder, baseline retrieval framework, and further settings. The two models that were fine-tuned on GMDB sometimes had different values, these models are indicated as m0 and m1.}\label{supp:tbl:implementation}
\begin{tabular}{@{}l p{.63\linewidth}@{}}
\toprule
Component & Setting \\
\midrule
Dataset version & GMDB v1.1.4 \\
Training data & GMDB-Freq training/validation split \\
Primary gallery setting & Unified GMDB-Freq+Rare gallery \\
Loss function & Class-weighted cross-entropy loss (scaled to [0.5-1.0]) \\
Model ensemble & [m0: iResNet50 GM-Arc, m1: iResNet100 GM-Arc, m2: iResNet100 ArcFace trained on GLINT360K] \\
Face detection and alignment & RetinaFace, aligned to the ArcFace five-landmark reference \\
Encoder input & $112 \times 112$ px aligned face crop \\
Test-time augmentation & [unaugmented, grayscale, horizontal flip, grayscale+horizontal flip] \\
Image representations & 12 model/TTA-specific embeddings per image \\
Embedding normalization & No explicit normalization applied to image embeddings, averaged patient embeddings, or disorder centroids \\
Distance metric & Cosine distance \\
Baseline disorder scoring & First occurrence of each disorder in image-level ranking; $k=1$ nearest neighbor disorder retrieval \\
Classifier-head learning rate & m0: 1e-3, m1: 1e-3 \\
Classifier-head weight decay & m0: 5e-4, m1: 5e-4 \\
Feature learning rate & m0: 5e-4, m1: 1e-3 \\
Feature weight decay & m0: 5e-5, m1: 0 \\
Batch size & m0: 64, m1: 128 \\
Optimizer & Adam optimizer \\
Learning rate scheduler & Reduce learning rate on plateau after 5 steps, with 5e-4 threshold, by factor of 0.5, with min. learning rate 1e-5 \\
Training epochs & 50, no early stopping or model selection \\
\bottomrule
\end{tabular}
\end{table}

\clearpage
\section{Analysis of inter- and intra-disorder similarity}\label{supp:sec:distances}

To characterize limitations of single-image nearest-neighbor retrieval in CFPS, we analyzed the distributions of intra-disorder (ground-truth) and inter-disorder (non-ground-truth) cosine distances over disorder-balanced samples of gallery-image pairs, splitting the intra-disorder distribution into frequent and rare disorder cohorts.

\begin{figure*}[pos=!ht]
  \centering
  \includegraphics[width=\linewidth]{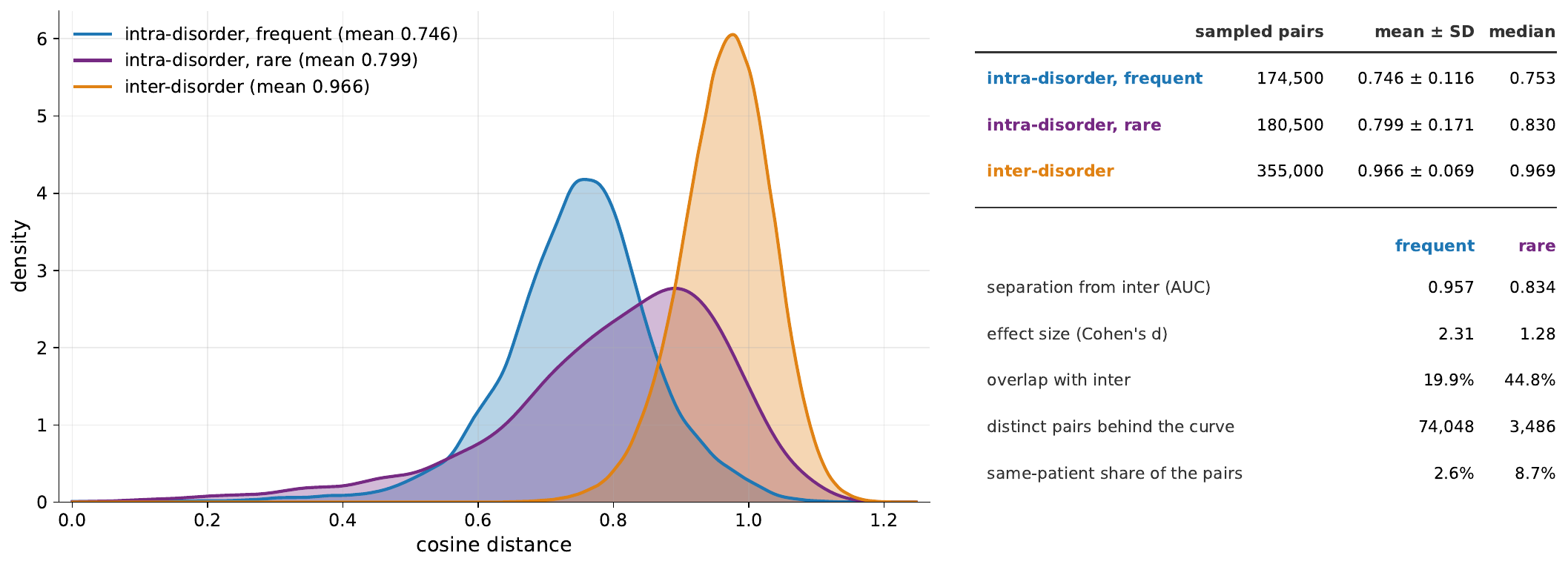}
  \caption{Intra-disorder cosine distance distributions for the frequent (blue) and rare (purple) cohorts, and the pooled inter-disorder distribution (orange), over disorder-balanced samples of unified-gallery image pairs. On the left, densities shown as Gaussian-smoothed histograms of cosine distance. On the right, the number of sampled pairs, mean, standard deviation, and median per group, and the area under the ROC curve separating intra- from inter-disorder distances (the probability that a randomly drawn inter-disorder pair is more distant than a randomly drawn intra-disorder pair), Cohen's $d$, the overlapping area with the inter-disorder density, the number of distinct pairs behind the sampled curve, and the same-patient share of the pairs.}
  \label{supp:fig:distances}
\end{figure*}

As shown in Supplementary Figure~\ref{supp:fig:distances}, the mean intra-disorder cosine distance is generally lower than the mean inter-disorder cosine distance for both cohorts, but markedly more so for GMDB-Freq disorders (0.746 versus 0.966) than for GMDB-Rare disorders (0.799 versus 0.966). Correspondingly, separation from the inter-disorder distribution is much weaker for rare disorders (AUC 0.834, Cohen's $d$ 1.28) than for frequent ones (AUC 0.957, Cohen's $d$ 2.31), and the rare intra-disorder density shares 44.8 \% of its area with the inter-disorder density, against 19.9 \% for the frequent cohort. The overlap between intra- and inter-disorder cosine distance distributions illustrates an intrinsic limitation of single-image $k=1$ nearest-neighbor retrieval, and this limitation is substantially more present for rare disorders. That means that for an individual test image, the nearest gallery image can belong to a different disorder even where the embedding space is well structured overall, and this is more than twice as likely for a patient with a rare disorder (unseen during training) as for one with a frequent disorder (seen during training).

\clearpage
\section{Growing number of patients with multiple images}\label{supp:sec:growth}

\begin{figure*}[pos=!ht]
  \centering
  \includegraphics[width=\linewidth]{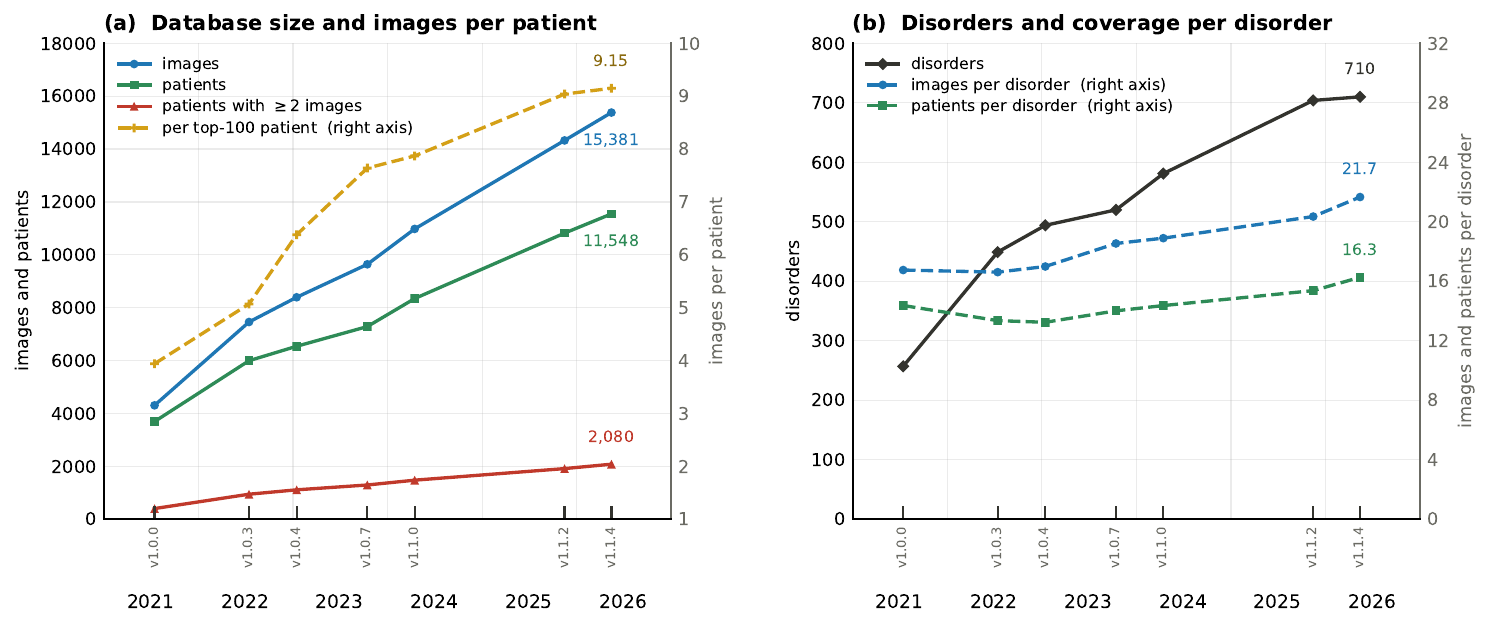}
  \caption{Major releases of GMDB from July 2021 (v1.0.0) to May 2026 (v1.1.4), plotted against release date. Solid lines are read on the left axis, dashed lines on the right axis. The number at the end of each line is its v1.1.4 value. (a) Left: total images (blue), total patients (green), and patients contributing at least two images (red). Right: mean images over the 100 most-represented patients of each release (yellow). (b) Left: number of distinct supported disorders (black). Right: mean images (blue) and mean patients (green) per disorder.}
  \label{supp:fig:growth}
\end{figure*}

Since the introduction of GestaltMatcher and the GestaltMatcher Database (GMDB) \citep{hsieh2022}, GMDB has grown steadily in the number of images, patients, and supported disorders \citep{lesmann2024}. To assess the relevance of patient-weighted disorder centroids, we quantified the number of images per patient across major GMDB versions. Although most patients are currently represented by a single image, the number of patients with multiple images has increased with database growth. Further, the number of patients with a disproportionate amount of images has also grown over time. Supplementary Figure~\ref{supp:fig:growth} shows these trends over time, as well as the total number of images, and patients. It further shows the mean number of images and patients per disorder. This trend motivates patient-weighted disorder representations, which prevent patients with many available images from dominating disorder centroids.

\clearpage
\section{Weighted hybrid individual-centroid scoring}\label{supp:sec:weighting}

\subsection{Initial weight selection}\label{supp:sec:weight-selection}

We performed the initial weight selection based on an analysis of the hybrid individual-centroid scoring on the GMDB validation set. The patients and images of this set were unseen during training, but it contains only disorders from the GMDB-Freq subset. As such, the selected $\lambda$ may not be ideal for all evaluation subsets. This validation set was therefore used as the only available held-out development set that did not require selecting hyperparameters on the evaluation subsets.

\begin{figure}[pos=!ht]
  \centering
  \includegraphics[width=.6\linewidth]{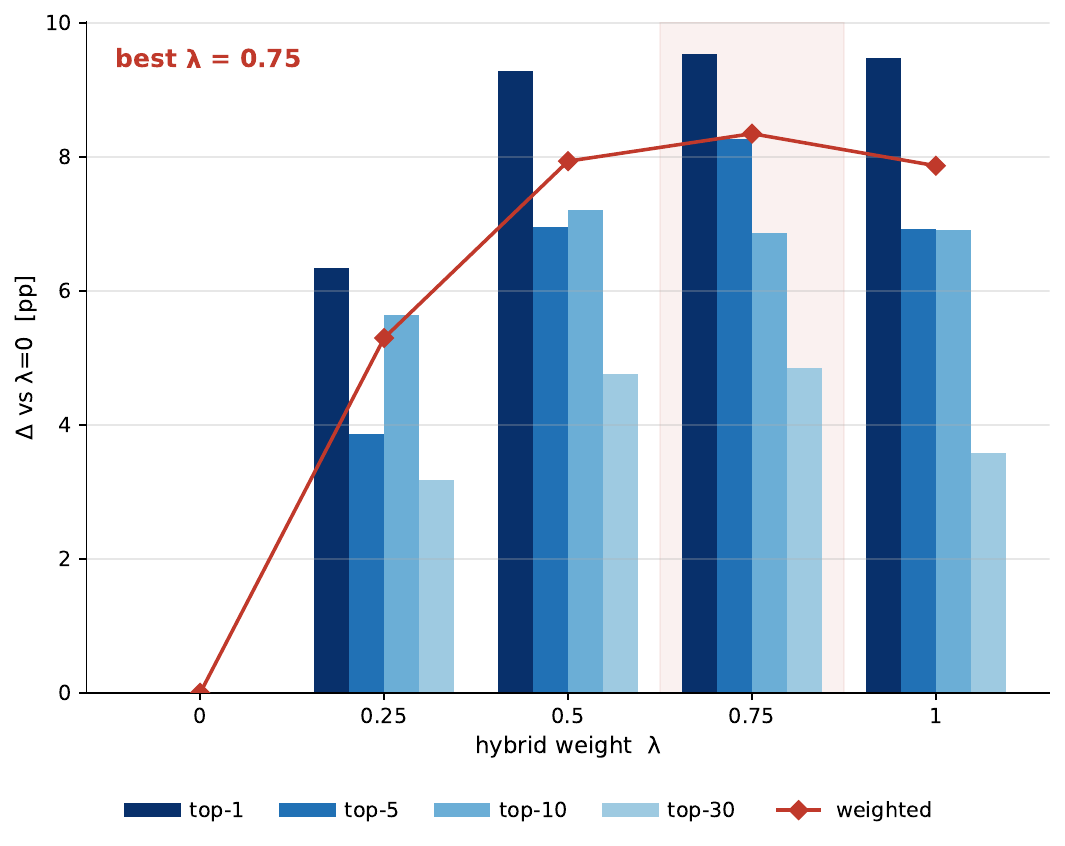}
  \caption{Improvement in top-1, top-5, top-10 and top-30 accuracy over $\lambda = 0$ (plain nearest-neighbor scoring), in percentage points, on the GMDB validation set for different values of $\lambda$. Each validation image is scored individually against the unified gallery with the validation images removed, using patient-weighted disorder centroids. The red line combines the four top-$N$ scores weighted by rank importance: top-1 is most important and weighted at 0.5, top-5 at 0.25, top-10 at 0.15 and top-30 at 0.1.}
  \label{supp:fig:weight-selection}
\end{figure}

Because the validation set contains 1,449 images of 1,080 patients with only disorders from the GMDB-Freq subset, it may favor weighting choices suited to well-represented disorders, such as those seen during training. We therefore selected $\lambda$ over a coarse grid, $\lambda \in \{0.00, 0.25, 0.50, 0.75, 1.00\}$. For this initial weight selection, we used a weighted validation objective combining top-1, top-5, top-10, and top-30 mean per-disorder accuracy. The main manuscript focuses on top-1, top-5, and top-10 accuracy as the primary retrieval cut-offs; top-30 was included only in this development step, with a low weight, to account for broader candidate-list prioritization while keeping the objective dominated by early-rank performance. Supplementary Figure~\ref{supp:fig:weight-selection} shows the results of this initial weight selection.

After our initial analysis, $\lambda=0.75$ achieved the highest combined score based on weighting the rank importance. It was also the best value for top-1 and top-30, and most clearly for top-5 on their own, whereas top-10 slighlty preferred $\lambda=0.5$. Therefore, we selected
$\lambda=0.75$ as fixed weight for hybrid individual-centroid scoring in the main experiments.

\clearpage
\subsection{Post hoc sensitivity analysis}\label{supp:sec:sensitivity}

We performed a post hoc sensitivity analysis over $\lambda \in \{0.00, 0.01, \ldots, 1.00\}$ to assess how strongly performance depended on the relative weighting of local nearest-neighbor and centroid evidence.
Because the analysis was performed on the evaluation subsets, it was used only to assess robustness to the hybrid weighting parameter and not to select the primary configuration. The main experiments use $\lambda=0.75$ throughout, as concluded from the initial weight selection on the validation set.

\begin{figure*}[pos=!ht]
  \centering
  \includegraphics[width=0.8\linewidth]{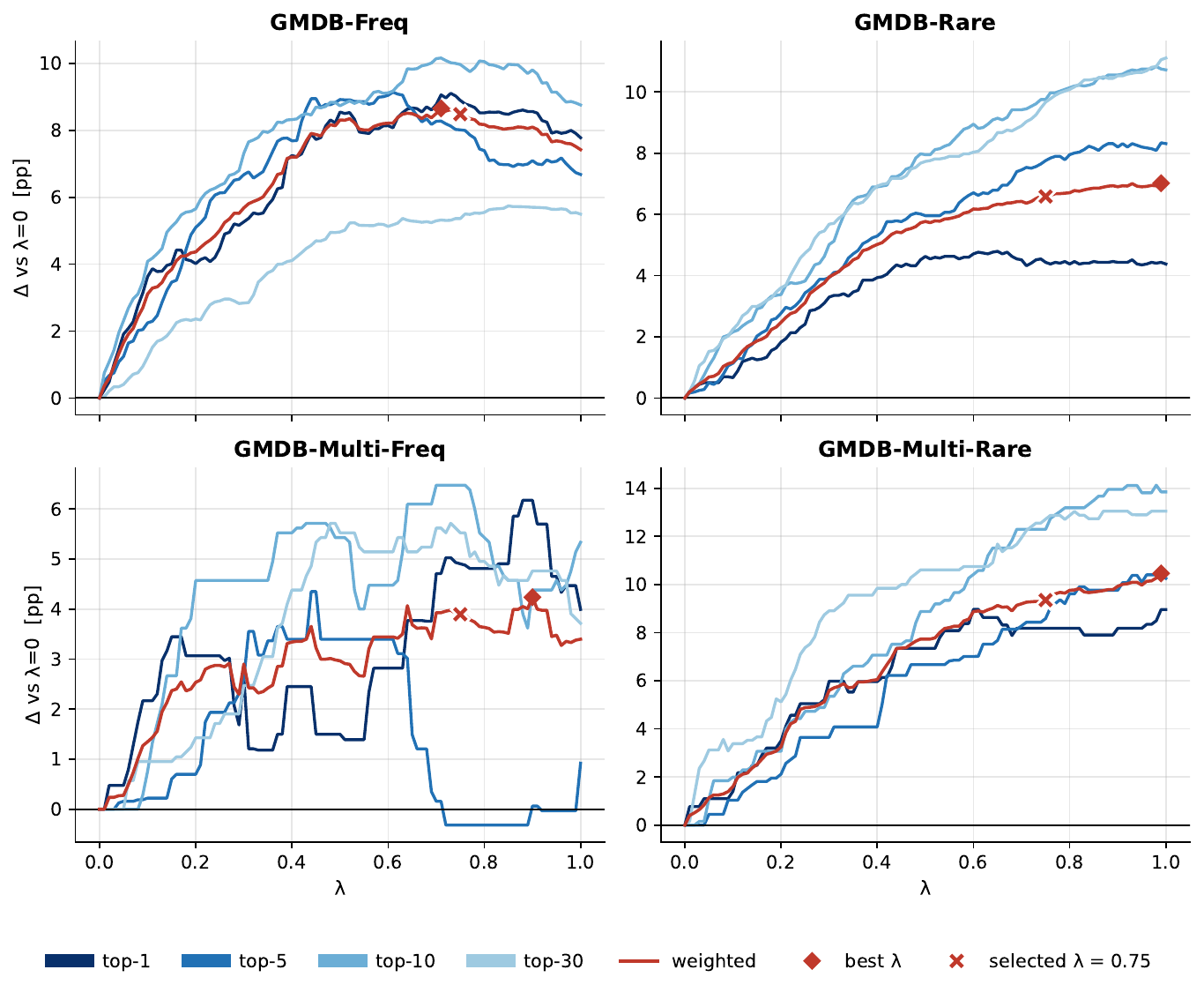}
  \caption{Improvement in top-1, top-5, top-10 and top-30 accuracy over $\lambda = 0$ (plain nearest-neighbor scoring), in percentage points, for different values of $\lambda$ on the four GMDB evaluation subsets, using patient-weighted disorder centroids with embedding-level fusion against the unified gallery. The red curve combines the four cut-offs weighted by rank importance (top-1 at 0.5, top-5 at 0.25, top-10 at 0.15, top-30 at 0.1). The diamond marks its maximum for that subset and the cross marks $\lambda = 0.75$, the value selected on the validation set.}
  \label{supp:fig:sensitivity}
\end{figure*}

As shown in Supplementary Figure~\ref{supp:fig:sensitivity}, the subsets differ in optimal $\lambda$. On GMDB-Freq, the interior optimum is reached at $\lambda = 0.71$ and declines beyond it, whereas both rare subsets improve monotonically and are therefore best served by centroid scoring alone. All four curves form a broad plateau above around $\lambda = 0.6$ and vary by at most 1.6 percentage points across it. Measured against each subset's own optimum, the $\lambda = 0.75$ selected on the validation set costs at most 0.43 percentage points on Freq, Rare and Multi-Freq, and 1.11 points on Multi-Rare. The Multi-Freq curve is visibly noisy because that subset contains only 156 patients.

This pattern is consistent with the long-tailed representation structure of GMDB, where disorders differ substantially in the number of available images and patients \citep{hsieh2022,hustinx2023,lesmann2024}, and with broader observations in long-tailed visual recognition that class-imbalance can affect recognition performance, particularly for sparsely represented classes \citep{zhang2023}. GMDB-Rare disorders are represented by few gallery images each, so any single nearest neighbor is potentially a noisy first selection, and averaging over a disorder's images can stabilize the score. GMDB-Freq disorders are represented by more gallery images spanning a wider range of facial presentations, so the disorder centroid may be pulled away from any particular presentation and a close individual match can carry information that the centroid might discard, which hybrid scoring partly retains.

\clearpage
\section{Multi-image performance stratified by number of test images}\label{supp:sec:num-images}

The main evaluation pools all multi-image test patients into a single number, which confounds how many images a patient has with how many are actually used. Patients with different numbers of images form different cohorts, so accuracy rising with the number of available images does not by itself show a benefit of combining them. To separate the two, we formed nested cohorts of patients with at least two, three, or four available images and scored each fixed cohort by using an increasing number of randomly sampled images per patient, drawing 50 random image subsets per patient at each number and averaging over them. Because the cohort is held constant along a curve, any change is attributable to the number of images used. The results of these multi-image stratified experiments are shown in Supplementary Figure~\ref{supp:fig:num-images}.

\begin{figure*}[pos=!ht]
  \centering
  \includegraphics[width=\linewidth]{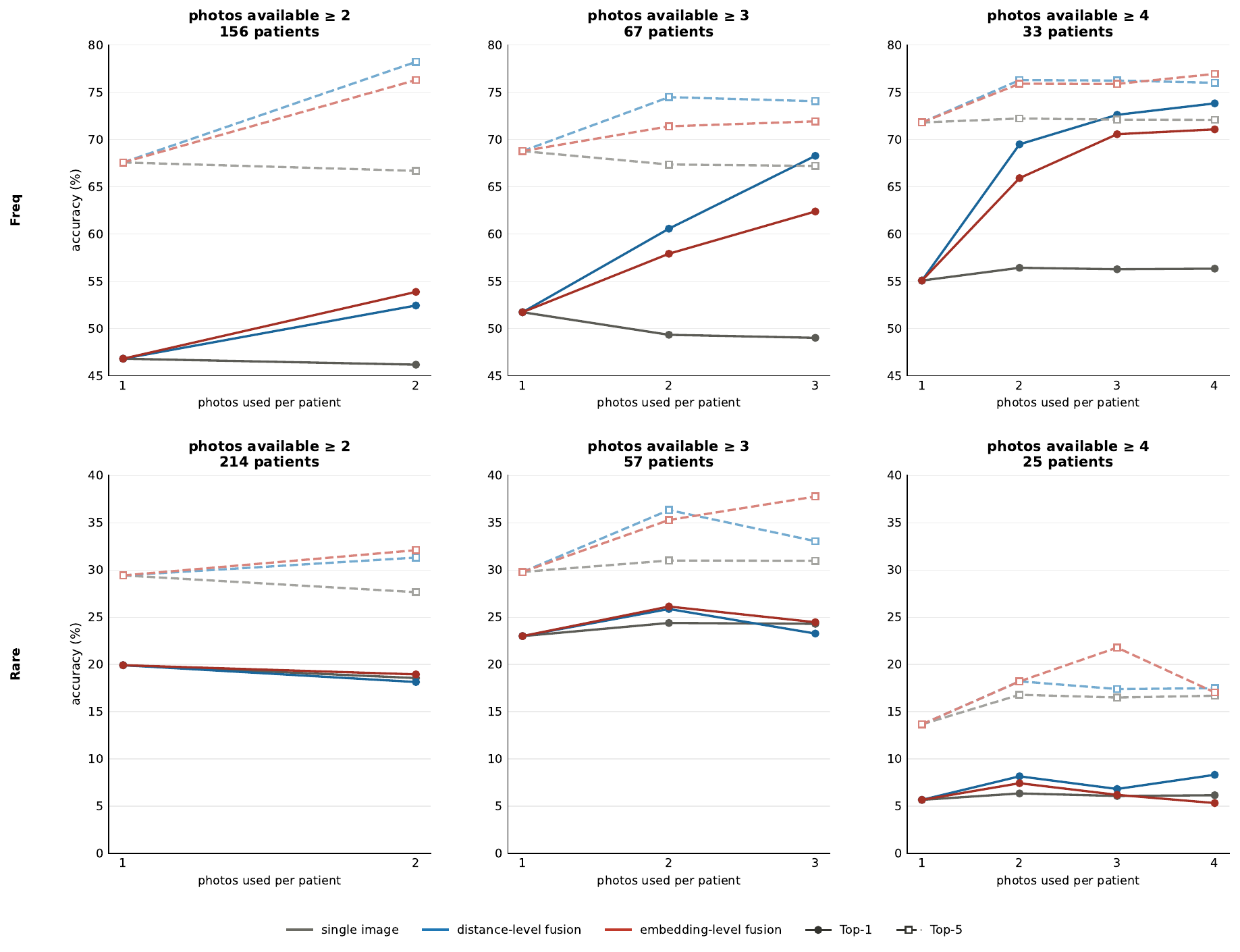}
  \caption{Retrieval performance as a function of the number of test images used per patient, within nested cohorts of patients having at least two, three, or four available images (left to right). The top row shows the GMDB-Frequent evaluation set and the bottom row the GMDB-Rare evaluation set, both evaluated against the unified gallery. The number of contributing patients is given in each panel title. Line colors indicate the aggregation approach: Single-image baseline (gray), distance-level fusion (blue), and embedding-level fusion (red). Solid lines denote top-1 and dashed lines top-5 accuracy.}
  \label{supp:fig:num-images}
\end{figure*}

On the GMDB-Multi-Freq evaluation set, both aggregation strategies improve steadily with the number of images used, while the single-image control stays flat in every cohort as expected. The largest gain comes from the first additional image, with some diminishing returns for the third and fourth, and top-5 accuracy saturates earlier than top-1. In the largest cohort, adding a second image improves top-1 accuracy by roughly 6 to 7 percentage points for both fusion strategies. On the GMDB-Multi-Rare evaluation set, no benefit is visible at top-1, but the decline is the same that is observed at the single-image control and thus a reflection of the sampled cohort. The benefit appears most prominently at the top-5 cut-off, where both fusion strategies gain while the control falls.

Overall, the benefit of patient-level aggregation observed in the main analysis was driven primarily by the GMDB-Multi-Freq evaluation set, while on GMDB-Multi-Rare disorders the benefit was confined to higher top-$N$ cut-offs and additional images of the same patient did not fully compensate for the limited gallery representations.

\clearpage
\section{Performance comparison of patient-level aggregation with oracle reference}\label{supp:sec:oracle}

To further interpret patient-level aggregation, we compared our proposed aggregation approaches with a retrospective best-single-image selection oracle reference. For each multi-image test patient, this reference selects the individual image with the best true-disorder rank.

Supplementary Table~\ref{supp:tbl:oracle} contrasts the two fusion strategies with the single-image baseline and with the oracle reference, for nearest-neighbor retrieval on multi-image test patients. On GMDB-Freq, both fusion strategies close a substantial part of the interval between the baseline and the oracle. On GMDB-Rare, they improve over the baseline from the top-5 cut-off onwards, while at top-1 the three remain comparable. Neither strategy reaches the oracle at any cut-off. The same holds for every operator and aggregation strategy we evaluated, including the proposed hybrid individual-centroid method.

In contrast to the retrospective optimal image-selection, aggregation does not choose among a patient's images. Rather, it combines the evidence they provide. Embedding-level fusion places the patient at a new position in the embedding space, at which image-specific variation such as head pose, expression or image quality may be partly canceled. Therefore, the fused representation can lie closer to the true-disorder region than any of the contributing images. Distance-level fusion rewards agreement across images so that a competing disorder favored by only one image can be displaced.

\begin{table}[width=\linewidth,cols=5,pos=!ht]
\caption{Top-$N$ accuracy (\%) for multi-image test patients, matched against the unified gallery with nearest-neighbor retrieval. The oracle reference retains, for each patient, only the image that ranks the correct disorder best.}\label{supp:tbl:oracle}
\begin{tabular*}{\tblwidth}{@{} LLRRR@{} }
\toprule
Approach & Evaluation set & Top-1 & Top-5 & Top-10 \\
\midrule
Single-image baseline & GMDB-Multi-Freq & 46.12 & 66.61 & 74.21 \\
Distance-level fusion & & 55.95 & 77.94 & 80.63 \\
Embedding-level fusion & & 56.04 & 76.29 & 77.97 \\
\textbf{Oracle optimal image-selection} & & \textbf{60.23} & \textbf{81.49} & \textbf{86.44} \\
Single-image baseline & GMDB-Multi-Rare & 18.54 & 27.62 & 32.58 \\
Distance-level fusion & & 17.75 & 30.43 & 36.24 \\
Embedding-level fusion & & 18.53 & 32.10 & 36.33 \\
\textbf{Oracle optimal image-selection} & & \textbf{26.16} & \textbf{36.39} & \textbf{41.53} \\
\bottomrule
\end{tabular*}
\end{table}

\begin{figure}[pos=p]
  \centering
  \includegraphics[width=\linewidth]{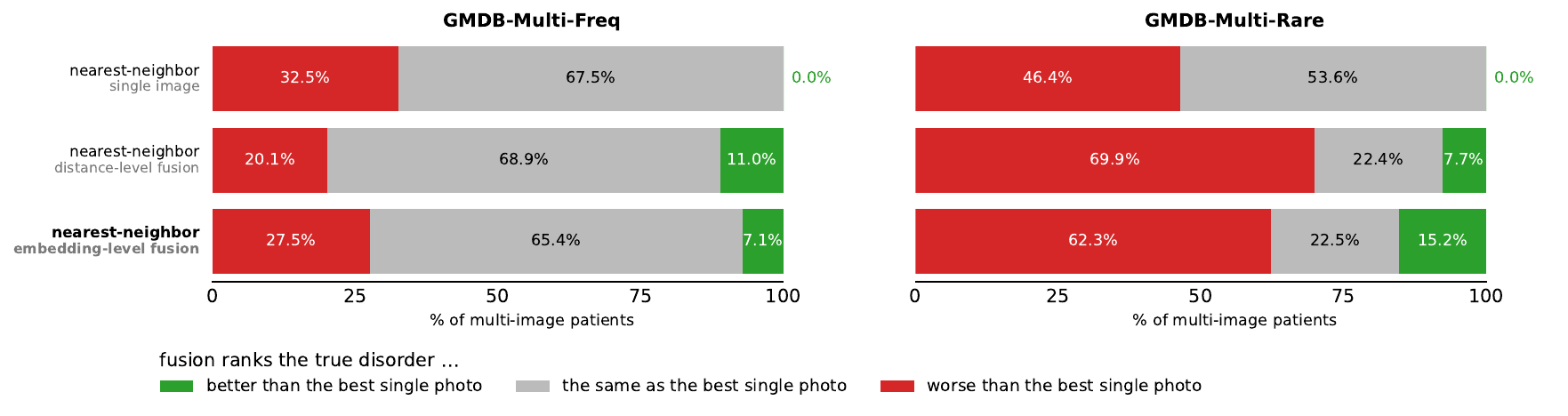}
  \caption{Per-patient comparison against the best available image, for multi-image test patients matched against the unified gallery with nearest-neighbor retrieval. Each bar shows the proportion of patients for whom the strategy ranks the correct disorder better than, the same as, or worse than the patient's best individual image. For the single-image baseline the proportions are the expectation over one randomly drawn image of the patient, which by definition can never rank better.}
  \label{supp:fig:oracle}
\end{figure}

Supplementary Figure~\ref{supp:fig:oracle} examines these notions per patient, comparing each method against the best image of the same patient. Both fusion methods manage to rank the correct disorder better than every individual image of a patient in between about 7 \% and 15 \% of cases, depending on the evaluation set and aggregation approach. The single-image baseline cannot improve performance in any way as it can, at best, match the patient's best performing image. The observed gains are outweighed by the patients for whom fusion falls behind their best image, which is why the mean remains below the oracle in Supplementary Table~\ref{supp:tbl:oracle}.

Although the oracle reference achieved the highest average performance overall, both embedding-level and distance-level fusion outperformed it for a subset of patients. This shows that patient-level fusion does not simply approximate selection of the best available image. Instead, fusion changes the full disorder ranking, including the ranks of incorrect candidate disorders. Thus, the oracle reference should be interpreted as a bound on retrospective single-image selection, not as a mathematical upper bound on patient-level aggregation.

\clearpage
\section{Centroid-based method performance stratified by gallery disorder size}\label{supp:sec:gallery-size}

To further explore the influence of using disorder centroids, we stratified the test patients by the number of gallery images belonging to their true disorder. Generally, we expected that the benefit of representing a disorder by a centroid instead of its individual gallery images is not constant across the gallery. When a disorder is represented by many images, the nearest-neighbor scoring already has enough options to find a similar reference image, so reducing the number of reference images by using centroids might not affect performance similarly to less represented disorders. When it is represented by only a few, the centroid mainly reduces the influence of individual atypical images of inter- and intra-disorder gallery patients. The same reasoning applies to hybrid individual-centroid scoring, which combines both benefits of the nearest-neighbor and centroid distance. The results of these stratified experiments is shown in Supplementary Figure~\ref{supp:fig:gallery-size}.

\begin{figure*}[pos=!ht]
  \centering
  \includegraphics[width=\linewidth]{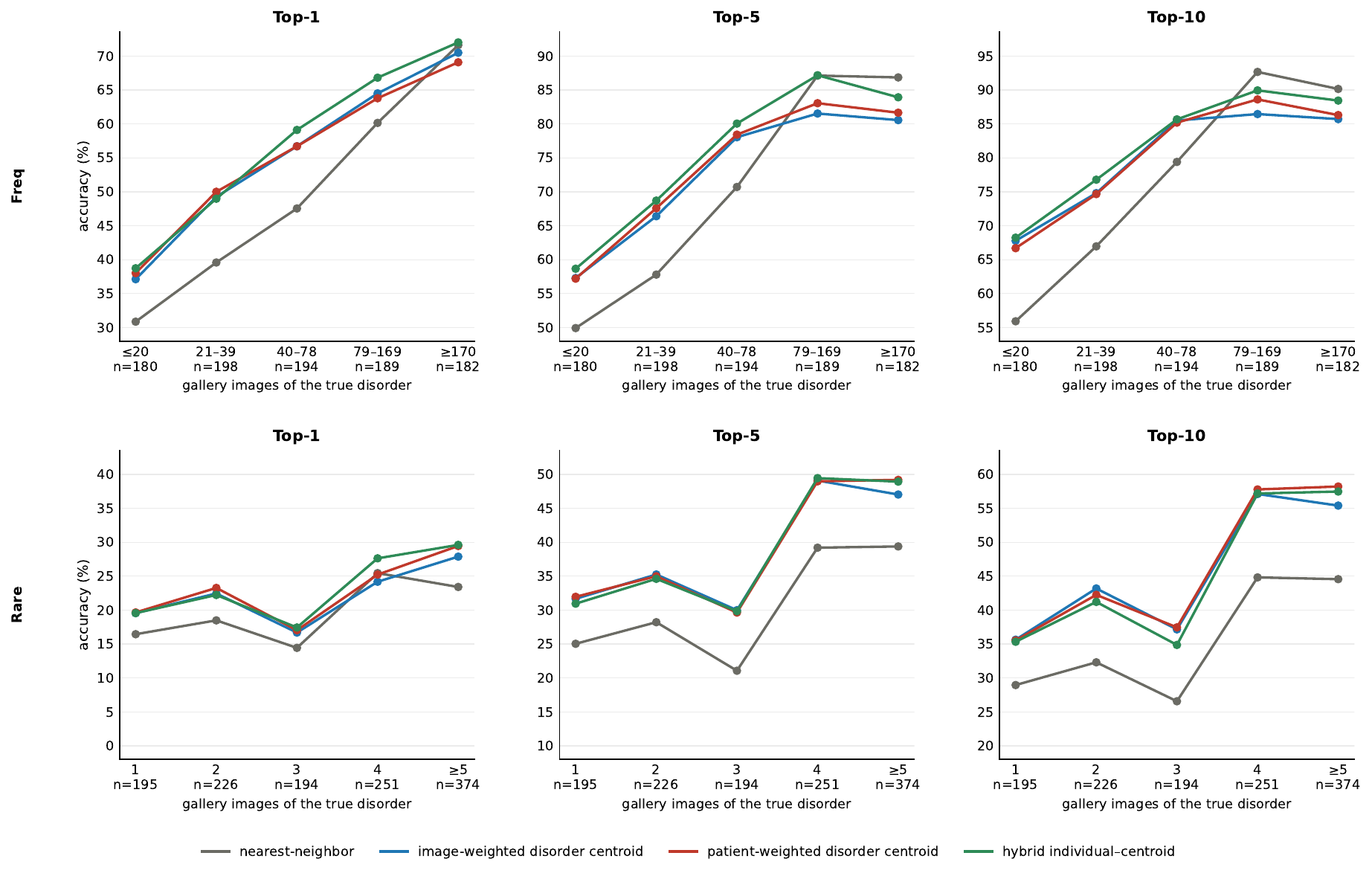}
  \caption{Accuracy of four retrieval approaches, stratified by the number of gallery images of the patient's true disorder. The top row shows GMDB-Freq and the bottom row GMDB-Rare, both evaluated against the unified gallery, with columns for top-1, top-5, and top-10 accuracy. Baseline image-level retrieval, image-weighted disorder centroids, patient-weighted disorder centroids, and hybrid individual-centroid scoring with $\lambda = 0.75$ are plotted in gray, blue, red, and green, respectively. The bins are defined separately per evaluation set, since a disorder from GMDB-Freq holds far more gallery images than a disorder from GMDB-Rare, and were chosen such that the set of test patients is split in quintiles as equally as possible. The number of contributing test patients is given below each bin.}
  \label{supp:fig:gallery-size}
\end{figure*}

Accuracy increases markedly with the number of available gallery images in both evaluation sets, which reflects that well-represented disorders are easier to retrieve and not an effect of the aggregation itself. Within each group, both centroid variants outperform baseline image-level retrieval as long as the disorder is sparsely represented. On GMDB-Freq this advantage is largest for disorders with fewer than about 80 gallery images, decreases as the gallery grows, and reverses for the best-represented disorders, where the baseline is the strongest approach at top-5 and top-10. On GMDB-Rare, the centroid approaches are superior in all groups, with the largest margins at top-5 and top-10. The non-monotonic shape of these curves reflects the small number of patients per group. Image-weighted and patient-weighted centroids are nearly indistinguishable over the entire range, with a slight advantage of patient-weighting for the most strongly represented disorders at top-5 and top-10.

Hybrid individual-centroid scoring follows the centroid approaches closely where the gallery is sparse, and separates from them toward the well-represented end of GMDB-Freq. It is most visible at top-5 and top-10, where the centroid approaches fall below the baseline for the largest disorders while the hybrid recovers most of the gap, thus recovering the individual-image evidence that centroid-only scoring discards.

Overall, the aggregation gains reported in Table~3 in Section~4.2 originate from sparsely represented disorders, whereas the additional gain of hybrid scoring in Table~3 in Section~4.3 originates from well represented ones. This also explains why hybrid scoring improves on GMDB-Freq but does not improve upon centroid scoring on GMDB-Rare, since unseen rare disorders lie entirely within the sparsely represented regime, in which individual-neighbor evidence appears less reliable when only few gallery cases are available.

\clearpage
\section{Detailed component-wise ablation study}\label{supp:sec:ablation}

\subsection{Ablations in unified-gallery setting}\label{supp:sec:ablation-unified}

A full ablation study of all components was performed on the GMDB evaluation sets using the unified gallery set to get a complete picture of the complementary nature of each component. These ablations are shown in Supplementary Figure~\ref{supp:fig:ablation-unified}.

\begin{figure*}[pos=!ht]
  \centering
  \includegraphics[width=.8\linewidth]{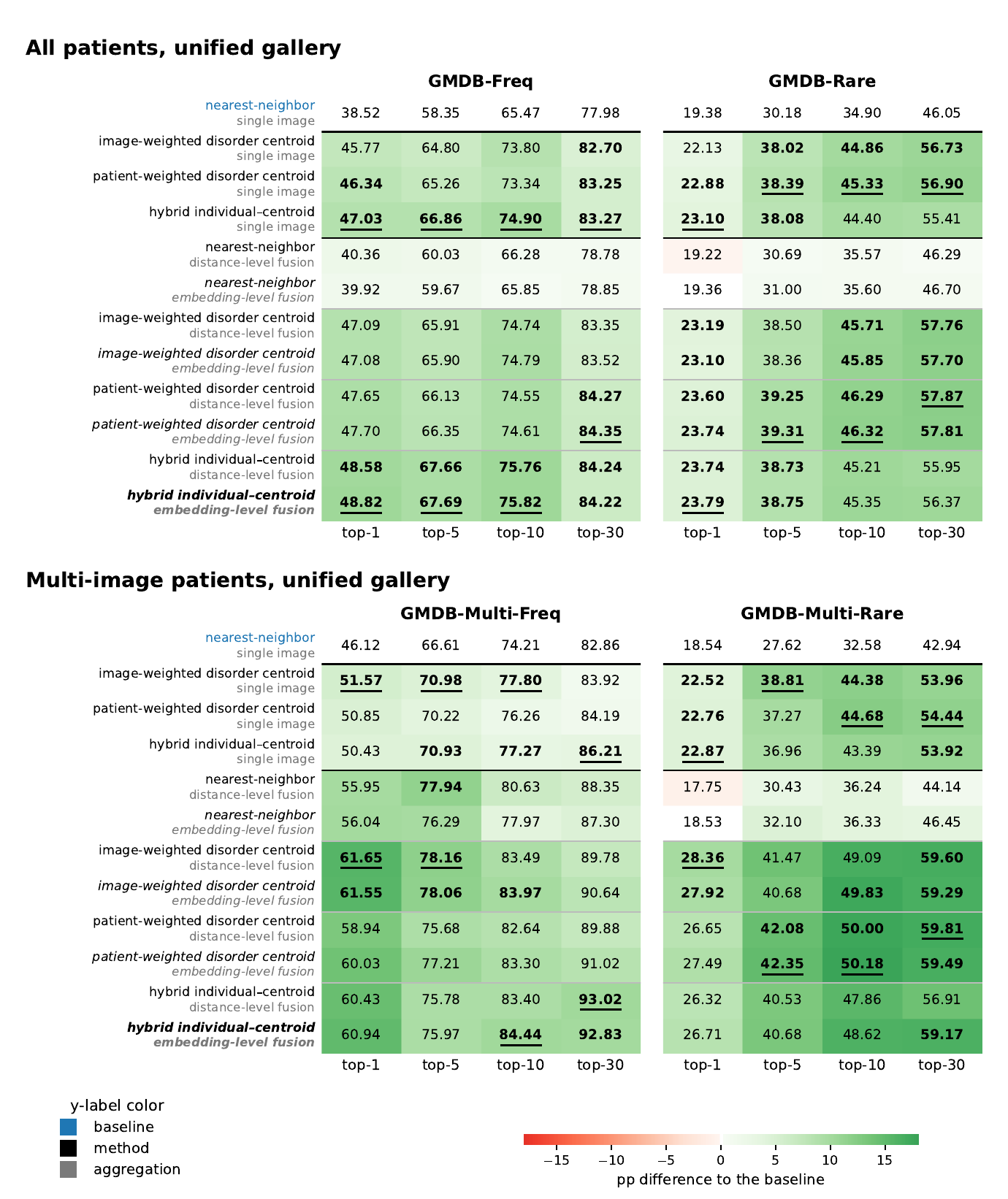}
  \caption{Component-wise ablation on the unified gallery. Top-$N$ accuracy (\%) for 12 methods (rows) on the all-patient sets (upper section) and their multi-image subsets (lower section). The horizontal line separates single-image from multi-image configurations, and the first row of each panel is the nearest-neighbor single-image baseline (blue label). Color encodes the difference in percentage points to that baseline. Underline marks the best configuration within its half of the panel, bold face those within 0.7 pp of it, and the bold row label the final configuration.}
  \label{supp:fig:ablation-unified}
\end{figure*}

The ablation shows that several aggregation components provide overlapping rather than purely additive gains. For example, patient-weighted centroids improved centroid-only retrieval on the all-patient sets, while on the multi-image subsets the image-weighted centroid was ahead at top-1. Even though hybrid individual-centroid scoring was best on the all-patient sets, pure centroid scoring performed similarly or better on the multi-image subsets. The level at which the images of a patient are combined (distance-level or embedding-level), had only a minor influence throughout. The final configuration aimed at balanced performance across datasets and methodological robustness, rather than subset-specific optimization.

\clearpage
\subsection{Ablations in split-gallery setting}\label{supp:sec:ablation-split}

To support backward comparability with earlier works using the GMDB, the ablations have also been performed using the split-gallery settings. The resulting ablations are shown in Supplementary Figure~\ref{supp:fig:ablation-split}. In the split-gallery settings, the gain of the final method over the baseline is similar to the change observed in the unified-gallery setting.

\begin{figure*}[pos=!ht]
  \centering
  \includegraphics[width=.8\linewidth]{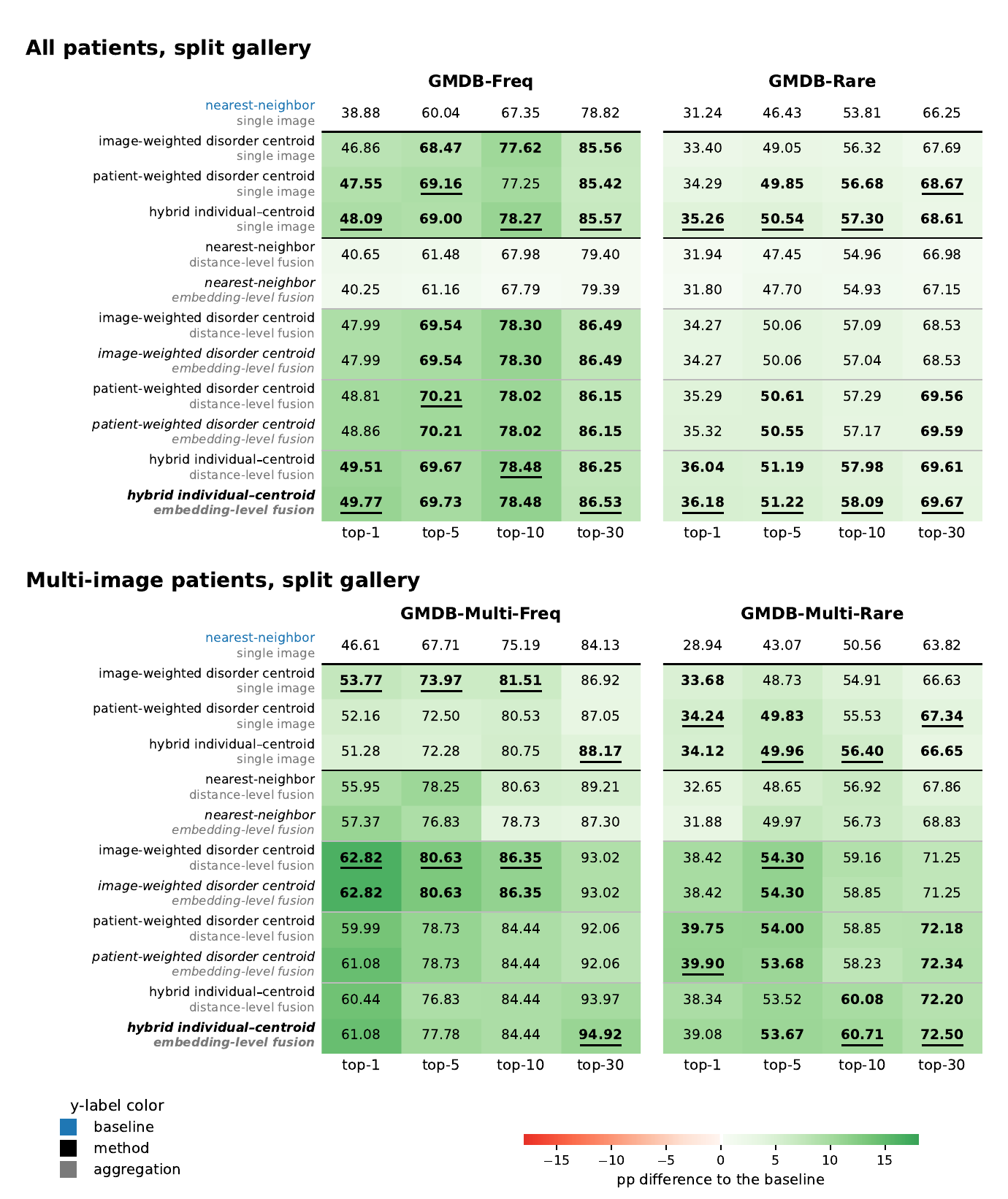}
  \caption{The same ablation against the split-galleries. Layout, configurations, metric and color scale are the same as in Supplementary Figure~\ref{supp:fig:ablation-unified}.}
  \label{supp:fig:ablation-split}
\end{figure*}

\clearpage
\section{Preliminary analysis of confounders on embedding-level fusion for patient aggregation}\label{supp:sec:confounders}

We analyzed the influence of confounders for test patients with multiple images on the predictive performance. Since the number of images per patient is limited and we are therefore limited in controlling this confounder analysis, these results are strictly preliminary.

For each test patient with at least two images, we quantified the benefit of embedding-level aggregation under nearest-neighbor retrieval as the top-30 rank gain. For this, we computed the averaged rank of the true disorder for a patient's individual images, minus the rank following embedding-level patient aggregation. All ranks are capped at 30 before averaging, such that movement outside the top-30 cut-off counts as no change. We then computed the Spearman rank correlation between this gain and seven per-patient variables: the number of images, the age spread across them, the smallest age gap between any of the patient's images and any gallery image of a patient with the true disorder, the mean age, sex, the number of gallery images of the true disorder, and the embedding norm, averaged over the patient's images. Each patient contributes a single observation, and confidence intervals were obtained by bootstrapping over patients.

\begin{figure}[pos=!ht]
  \centering
  \includegraphics[width=0.8\linewidth]{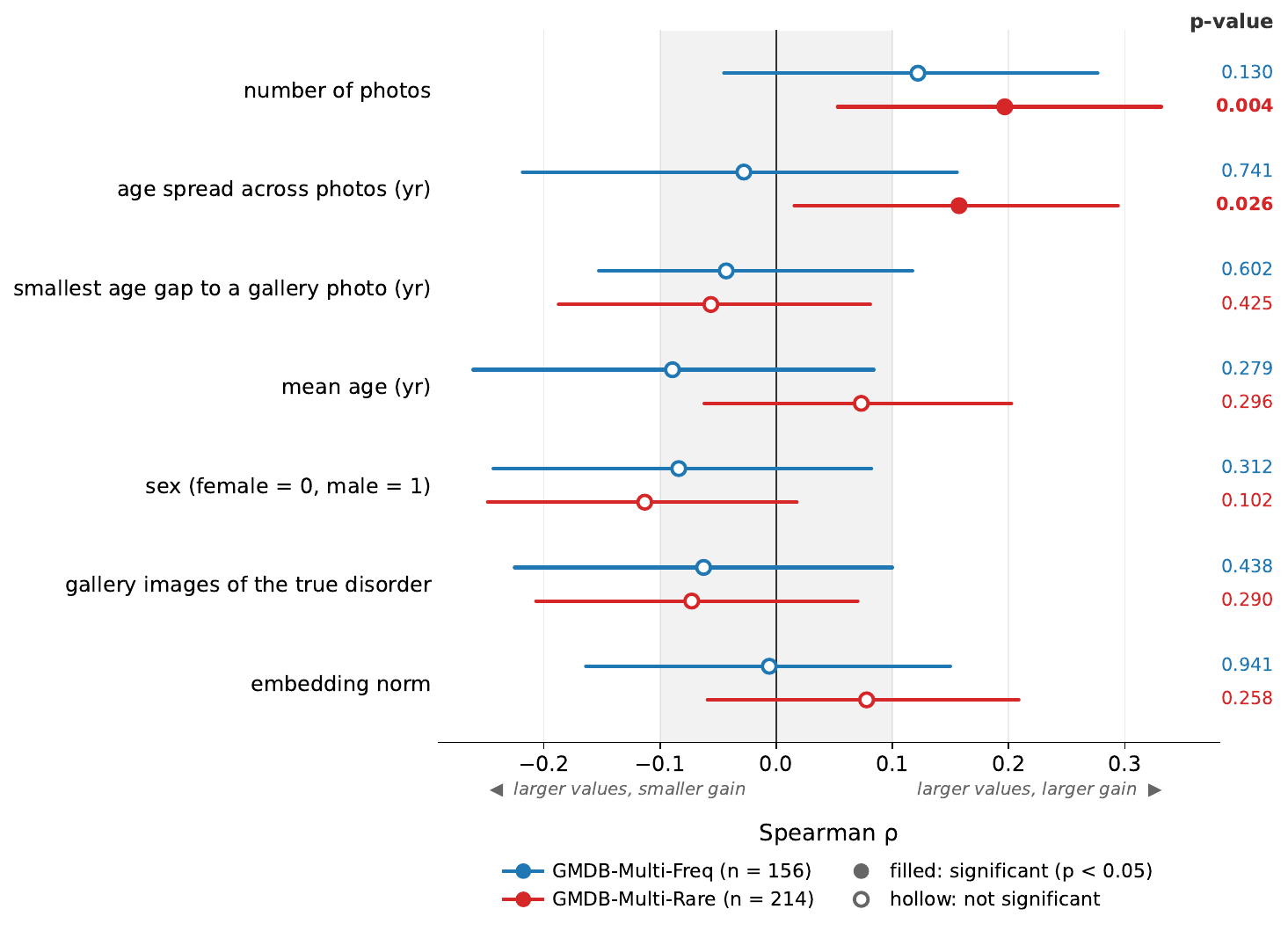}
  \caption{Spearman rank correlation between per-patient variables and the top-30 rank gain from embedding-level patient aggregation, for GMDB-Multi-Freq (blue) and GMDB-Multi-Rare (red). Points give the correlation, bars its 95 \% bootstrap confidence interval over patients, and the right-hand column the uncorrected $p$-value. Filled points are significant at $p < 0.05$. Positive correlations indicate that larger values of the variable correlate with a larger gain from aggregation, while negative correlations go with a smaller gain. The shaded band marks $|\rho| < 0.1$, below which, by convention, a correlation is not interpreted regardless of its $p$-value.}
  \label{supp:fig:confounders}
\end{figure}

A positive correlation means that patients with more of a given variable gain more from aggregation, while a negative correlation means that they gain less. Supplementary Figure~\ref{supp:fig:confounders} shows the results of these analyses.

In GMDB-Multi-Freq, no variable showed an association since every confidence interval covered zero, while in GMDB-Multi-Rare we found two correlations. Patients gained more from aggregation when contributing more images ($\rho = 0.20$) and when those images span a wider age range ($\rho = 0.16$). Because data is scarce and performance depends on both the test patients and the gallery set, these results should be treated as a direction to revisit once more multi-image patients are available, not as established associations.

\clearpage
\section{Proportional nearest-neighbor scoring}\label{supp:sec:propnn}

The baseline disorder ranking uses the first occurrence of each disorder in the image-level gallery ranking, corresponding to $k=1$ nearest-neighbor (NN) disorder scoring. While this approach is simple, it can be sensitive to individual gallery images. A single gallery image from a well-represented disorder may receive a small distance to the test image and therefore cause that disorder to be ranked highly, even if the remaining gallery images of the same disorder are less similar.

To reduce sensitivity to single-image matches, we evaluated proportional NN scoring. Instead of using only the nearest gallery image for each disorder, this strategy aggregates the distances of a disorder-specific number of nearest gallery images. For disorder $d$, let $M_d$ be the number of gallery images associated with that disorder and let $\alpha \in (0,1]$ be the proportion of gallery images used for scoring, so that $k_d = \lceil \alpha M_d \rceil$. The proportional NN distance is then computed as the mean distance to the $k_d$ nearest gallery images of disorder $d$:
\begin{equation}\label{supp:eq:propnn}
D_{\mathrm{propNN}}(t,d) = 1/k_d \sum_{g \in \mathcal{N}_{k_d}(t,d)} D(t, g),
\end{equation}
where $\mathcal{N}_{k_d}(t,d)$ is the set of $k_d$ nearest gallery images of disorder $d$, and $D(t,g)$ is the cosine distance averaged over the model / augmentation representations, as defined in the main text. Note that for $k_d=1$ we retrieve nearest-neighbor scoring, and for $\alpha=1$ we retrieve the mean-distance score over all of the disorder's gallery images. The latter is related to, but distinct from centroid scoring, which instead computes the distance to the disorder's mean embedding.
In different notation, proportional NN can be seen as $\alpha$\%-NN.

This strategy adapts the amount of local evidence used for scoring to the number of available gallery images per disorder. For disorders with many gallery images, the disorder score will depend on multiple nearest gallery images rather than a single nearest neighbor. Consequently, one unusually close gallery image might have less influence on the final ranking, unless it is supported by additional nearby images from the same disorder. For disorders with few gallery images, the ceiling keeps $k_d=1$ for every $\alpha \leq 1/M_d$, so ultra-rare disorders represented by only a small number of gallery images continue to be scored by their single nearest neighbor.
We evaluated proportional nearest-neighbor scoring as an exploratory alternative to the $k=1$ NN disorder score used in the baseline framework. This strategy improved retrieval especially in the unified-gallery setting, with the largest gains observed for GMDB-Rare. This suggests that requiring support from multiple nearby gallery images can reduce sensitivity to isolated gallery-image matches, particularly when images of patients with rare disorders are evaluated against a larger and more imbalanced gallery containing frequent disorders with many reference images.

\begin{figure*}[pos=p]
  \centering
  \includegraphics[width=\linewidth]{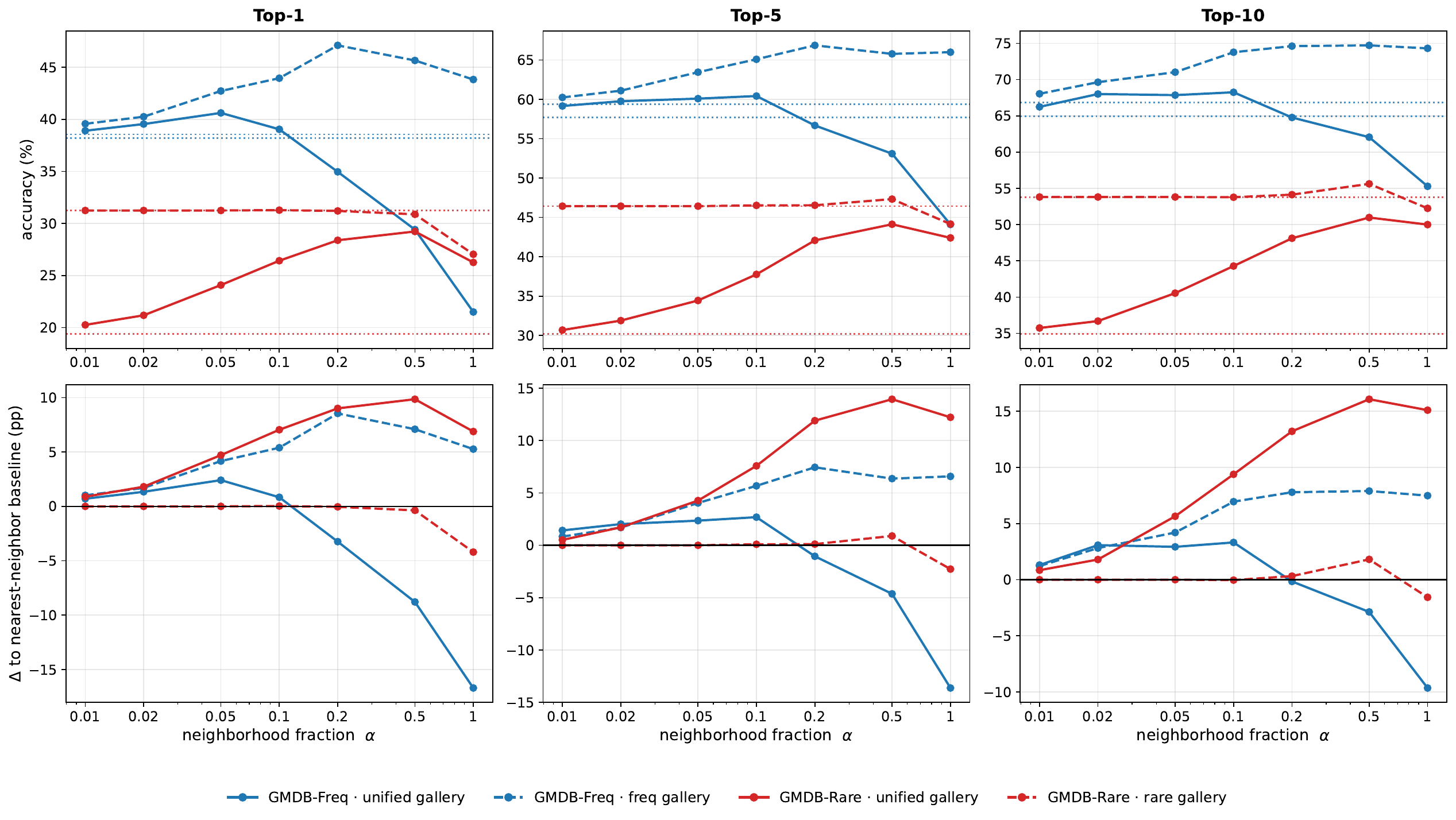}
  \caption{Resulting performance of proportional NN scoring with $\alpha$ as (logarithmic) x-axis, (top) per-disorder accuracy on y-axis, and (bottom) the difference to the $k=1$ nearest-neighbor baseline of the same curve, in percentage points, measured in per-disorder top-1, top-5, and top-10 accuracy. Line color indicates the evaluation set: GMDB-Freq in blue and GMDB-Rare in red, and line style the gallery, with the unified gallery (Freq+Rare) drawn solid and the evaluation set's own gallery dashed. Dotted horizontal lines mark the $k=1$ baseline of the corresponding curve.}
  \label{supp:fig:propnn}
\end{figure*}

Supplementary Figure~\ref{supp:fig:propnn} shows the impact of different proportional $\alpha$-values on the performance of the GMDB-Freq and GMDB-Rare evaluation sets, for both the standard split-galleries and unified-gallery.

As can be seen, when evaluating GMDB-Freq using its GMDB-Freq split-gallery, up to around $\alpha=0.2$ led to large performance improvements, while the GMDB-Rare split-gallery setting is mostly unaffected. The latter is likely due to $\lceil \alpha M_d \rceil = 1$ for most of its small galleries. However, when evaluating the unified gallery set (Freq+Rare), the two evaluation sets diverge beyond around $\alpha=0.1$. GMDB-Freq turns negative while GMDB-Rare keeps improving up to $\alpha=0.5$. Both sets improve simultaneously only in the range $\alpha=0.05$ to 0.1, with $\alpha=0.05$ leading to overall improvements in both sets. The contrast between the two gallery conditions follows from a de-biasing effect: raising $k_d$ for the frequent disorders helps rare queries, for which those disorders act as distractors, but penalizes frequent queries against rare disorders.

Although proportional NN scoring showed performance improvements by itself, the gains were substantially reduced when it was combined with hybrid individual-centroid scoring. On GMDB-Rare only a fraction of the improvement remained on top of the hybrid, and on GMDB-Freq the combination did not add anything on top of the hybrid improvements. A likely explanation is that the hybrid already aggregates all gallery images of a disorder through its centroid term, so the two mechanisms are largely redundant. In addition, proportional NN distances are generally larger than $k=1$ NN distances because they average over multiple gallery images, rather than assuming only the closest which may change the scale and balance of the hybrid distance. For this reason, proportional NN scoring was not included in the primary full aggregation framework.

\clearpage

\putbib
\end{bibunit}

\end{document}